%% file: neurips_2025.tex
\documentclass{article}

\usepackage[preprint]{neurips_2025}

\input{math_commands.tex}

\usepackage[utf8]{inputenc} 
\usepackage[T1]{fontenc}    
\usepackage{hyperref}       
\usepackage{url}            
\usepackage{booktabs}       
\usepackage{amsfonts}       
\usepackage{nicefrac}       
\usepackage{microtype}      
\usepackage{xcolor}         

\usepackage{bm}

\usepackage{microtype}
\usepackage{subfigure}
\usepackage{hyperref}
\usepackage{setspace}
\usepackage{bmpsize}
\usepackage{algorithm}
\usepackage{algpseudocode}
\definecolor{cvprblue}{rgb}{0.21,0.49,0.74}
\hypersetup{
    breaklinks=true,
    colorlinks=true,
    citecolor=cvprblue,
    linkcolor=purple,
    urlcolor=cyan,
}

\usepackage{amsmath}
\usepackage{amssymb}
\usepackage{mathtools}

\usepackage{amsthm}
\usepackage{verbatim}
\usepackage{multirow}
\usepackage{wrapfig}
\usepackage{float}
\usepackage[capitalize,noabbrev]{cleveref}

\theoremstyle{plain}
\newtheorem{theorem}{Theorem}[section]
\newtheorem{proposition}[theorem]{Proposition}
\newtheorem{lemma}[theorem]{Lemma}

\theoremstyle{definition}
\newtheorem{definition}[theorem]{Definition}
\newtheorem{assumption}[theorem]{Assumption}
\theoremstyle{remark}
\newtheorem{remark}[theorem]{Remark}
\usepackage[textsize=tiny]{todonotes}

\newcommand{\norm}[1]{\left\|#1\right\|}
\newcommand{\dotprod}[1]{\left\langle #1\right\rangle}
\newcommand{\EE}{\mathbb{E}}

\newcommand{\cL}{\mathcal{L}}
\newcommand{\RR}{\mathbb{R}}

\newcommand{\cB}{\mathcal{B}}
\newcommand{\ccV}{\mathcal{V}}
\title{FZOO: Fast Zeroth-Order Optimizer for Fine‑Tuning Large Language Models towards Adam‑Scale Speed}

\author{%
  Sizhe Dang\thanks{Equal contribution.}\\
  Xi’an Jiaotong University \\
  \And
  Yangyang Guo\footnotemark[1]\\
  Xi’an Jiaotong University\\
  \And
  Yanjun Zhao\footnotemark[1]\\
  Xi’an Jiaotong University \\
  \And
  Haishan Ye\footnotemark[1]~~\thanks{Corresponding author.}\\
  Xi’an Jiaotong University\\
  SGIT AI Lab\\
  \And
  Xiaodong Zheng \\
  Xi’an Jiaotong University \\
  \And
  Guang Dai \\
  SGIT AI Lab\\
  \And
  Ivor Tsang \\
  A*STAR\\
}

\newcommand{\name}{FZOO~}
\newcommand{\nameo}{FZOO}

\begin{document}

\maketitle
{\centering
\small
\texttt{\{darknight1118, no.314016, yanjun.zhao, zxd\_xjtu\}@stu.xjtu.edu.cn}\\
\texttt{yehaishan@xjtu.edu.cn, guang.dai@gmail.com, ivor\_tsang@cfar.a-star.edu.sg}
\par
}
\begin{abstract}
Fine-tuning large language models (LLMs) often faces GPU memory bottlenecks: the backward pass of first-order optimizers like Adam increases memory usage to more than 10 times the inference level (e.g., 633~GB for OPT-30B). Zeroth-order (ZO) optimizers avoid this cost by estimating gradients only from forward passes, yet existing methods like MeZO usually need tens of times more steps to converge. Can this trade-off between speed and memory in ZO be fundamentally improved? Normalized-SGD, for instance, demonstrates strong empirical performance with greater memory efficiency than Adam. In light of this, we introduce \textbf{FZOO}, a Fast Zeroth-Order Optimizer towards Adam-Scale Speed. On the one hand, FZOO reduces the total forward passes needed for convergence by employing batched one-sided estimates that adapt step-sizes based on the standard deviation of batch losses. On the other hand, it accelerates per-batch computation through the use of Rademacher random vector (±1) perturbations coupled with CUDA's parallel processing capabilities. 
Extensive experiments on diverse models (including RoBERTa-large, the OPT family (350M-66B), Phi-2, and Llama3) across 11 varied downstream tasks validate FZOO's effectiveness. On average, FZOO outperforms MeZO by +3\% in accuracy while requiring 3$\times$ fewer forward passes. Notably, for the RoBERTa-large model, FZOO achieves average improvements of +5.6\% in accuracy and 18$\times$ reduction in forward passes compared to MeZO, achieving convergence speeds comparable to Adam. We also provide theoretical analysis proving FZOO’s formal equivalence to a normalized-SGD update rule and establishing its convergence guarantees. Beyond full-parameter tuning, FZOO plugs smoothly into PEFT techniques, unlocking even larger memory savings. Taken together, our results make single-GPU, high-speed, full-parameter fine-tuning realistic today and point toward future work on memory-efficient pre-training. \textbf{Project Homepage}: \url{https://dkmiyan.github.io/FZOO_hp}

\end{abstract}

\section{Introduction}
\label{sec_introduction}

\begin{figure*}[h]
\centering
\includegraphics[width=1\linewidth]{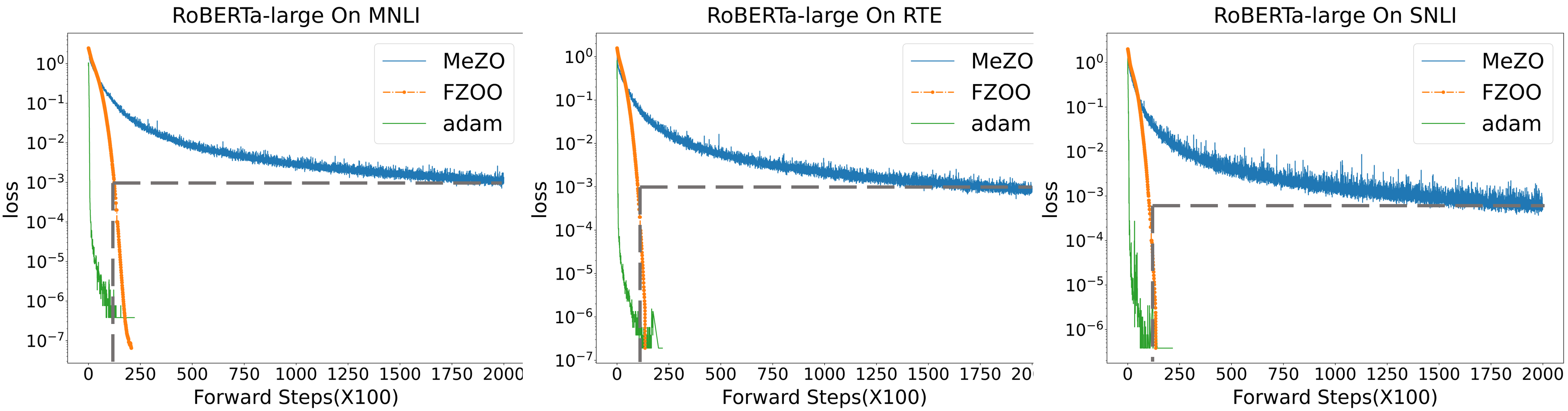}
\vspace{-7mm}
\caption{Performance of MeZO, Adam and FZOO on different tasks when fine-tuning RoBERTa-large model. For the sake of uniform comparison, we convert Adam's backward pass into 3 forward passes. FZOO can achieve 18$\times$ speedup compared with MeZO, nearly 20$\times$ that of Adam.}
\label{fig:fzoo_adam}
\vskip -0.02in
\end{figure*}

Large language models (LLMs) have become the workhorse of natural‑language and multi‑modal applications~\citep{BERT,gururangan-etal-2020-dont,ouyang2022training}, yet they often require additional fine‑tuning to excel on downstream tasks.  Current practice relies on first‑order optimizers such as Adam~\citep{Adam}, whose backward pass stores all activations and gradients, linearly multiplying the memory cost—fine‑tuning an OPT‑30B model then consumes \textbf{633 GB} of GPU memory, more than \textbf{10×} its inference memory cost.  This creates the \emph{memory wall} that now limits model training in resource‑constrained scenarios.

Two main strategies have been explored to overcome this wall.  \emph{Parameter‑efficient tuning} (PEFT)~\citep{hu2022lora,li-liang-2021-prefix, dettmers2023qlora, zhao2024galore, pan2024lisa} (e.g., LoRA~\citep{hu2022lora})  updates only a small subset of weights and thus lowers memory cost, but it still requires the backward pass and can lag full‑parameter tuning (FT) on some difficult tasks. The other approach turns to \emph{zeroth‑order} (ZO) optimization~\citep{mezo,hizoo}: methods such as MeZO~\citep{mezo} use multiple forward passes to replace the backward pass when estimating gradients. This brings the memory overhead of fine‑tuning down to the inference level, but this benefit comes with a significant trade-off in speed, as convergence is often about  \textbf{$20\times$} slower than Adam on RoBERTa-large (Figure \ref{fig:fzoo_adam}). \textbf{Does the use of zeroth-order methods always lead to a trade-off between speed and memory?}

We believe the answer is \textbf{no}. This is because ZO performance can be substantially enhanced in three key areas: (i) \textbf{Adaptive Step-Sizes.} While Adam uses costly momentum for adaptive steps, normalized-SGD adapts efficiently via gradient normalization without this overhead. Existing ZO methods, however, use inefficient fixed step-sizes for LLM fine-tuning. (ii) \textbf{Efficient Query Processing.} Furthermore, each ZO gradient estimate typically requires multiple forward queries. The overall speed of this multi-query process, however, is often hindered by both the computational expense of conventional perturbation sampling (e.g., Gaussian noise) and execution strategies that underutilize parallelism across these queries. (iii)\textbf{ Engineering Optimization.} Theoretically, one Adam step is twice as fast as a MeZO step (Adam’s backward pass takes roughly three times longer than a forward pass~\citep{alman2024fine}, making Adam equivalent to four forward passes versus MeZO’s two). Yet, due to insufficient forward-pass optimizations, MeZO’s actual per-step execution often exceeds Adam’s. Taken together, these observations suggest that a well‑engineered ZO optimizer could achieve \textbf{Adam‑level convergence speed while retaining inference‑level memory cost}.

To harness these opportunities for ZO enhancement, we introduce \textbf{FZOO}, a \textbf{F}ast \textbf{Z}eroth-\textbf{O}rder \textbf{O}ptimizer. FZOO achieves its significant speed-up through two primary strategies: First, it reduces total convergence passes by employing \emph{batched one-sided estimates}, where the standard deviation of batch losses informs an \emph{adaptive step-size rule}. Second, it accelerates each batch computation using efficient \emph{Rademacher random vector perturbations} ($\pm 1$) and \emph{CUDA-optimized parallelism} for the query batches.  Specifically, our contributions are: 

\begin{enumerate}
    \item \textbf{FZOO}, a novel ZO optimizer that uniquely adapts principles from normalized-SGD to the ZO regime. It is designed to achieve \emph{Adam-scale convergence speed} while operating at an \emph{inference-level memory usage}.
    \item We provide rigorous theoretical backing for FZOO, including a formal proof of its equivalence to a normalized-SGD update rule (Section~\ref{Equivalence2normalized-SGD}) and a comprehensive analysis establishing its \emph{convergence guarantees} under standard conditions (Section~\ref{sec_convergence}).
    \item Extensive FT experiments on diverse models (RoBERTa-large, OPT family 350M--66B, Phi-2, and Llama3) across 11 downstream tasks demonstrate FZOO's superior efficiency (Section~\ref{sec_experiments}). On average, FZOO surpasses the MeZO baseline by \textbf{+3\%} in accuracy with \textbf{3$\times$} fewer forward passes. Notably, as show in Fig.~\ref{fig:fzoo_adam} for the RoBERTa-large model it achieves average improvements of \textbf{+5.6\%} in accuracy and an \textbf{18$\times$} reduction in forward passes compared to MeZO, reaching effective convergence speeds \emph{comparable to Adam}.
    \item FZOO exhibits strong practical utility, smoothly integrating with PEFT techniques~\citep{dettmers2022gptint,dettmers2022bit} to unlock \emph{even larger memory savings} (Section \ref{orthogonality_hyperparameter}). Furthermore, it effectively optimizes both differentiable and non-differentiable objectives during FT (Section \ref{section_F1}).
\end{enumerate}

\section{Related Works}
\label{sec_related_works}

We provide a brief overview of first‑order, zeroth‑order and batch-based zeroth-order optimizers for large language models; see Appendix~\ref{app_related_works} for details.

\textbf{First‑order adaptive methods}~Standard LLM optimizers include SGD and its refinements~\citep{Adagrad} like Adam~\citep{Adam}, which uses memory-intensive momentum for its adaptive step-sizes. AdamW~\citep{adamw} improves generalization by refining weight decay. Inspiringly, methods like normalized-SGD~\citep{bernstein2018signsgd} also adapt steps effectively, often rivaling Adam's performance but crucially without this significant momentum overhead, offering a key insight for our work. However, all these first-order methods share a fundamental drawback: back-propagation's requirement to store activations and gradients incurs steep memory costs, severely bottlenecking the fine-tuning of large models.

\textbf{Zeroth‑order optimizers}~ZO methods sidestep back‑prop by estimating gradients with only forward passes. The classic SPSA estimator~\citep{ZO_GradientEstimator_SPSA} has powered applications from distributed control~\citep{distributed_ZO1,distributed_ZO2} to black‑box adversarial attacks~\citep{black_box_adversial_ZO1,black_box_adversial_ZO2,black_box_adversial_ZO3,black_box_adversial_ZO4}. MeZO~\citep{mezo} first showed that such forward‑only updates can fine‑tune LLMs to high accuracy while cutting GPU memory by several times (up to 12×). HiZOO~\citep{hizoo} augments MeZO with diagonal Hessian cues to accelerate convergence, but doing so doubles memory and introduces extra hyper‑parameters that must be hand‑tuned, trading one resource for another.  
A recent benchmark~\citep{zobenchmark} catalogues many other ZO estimators; nevertheless, none yet approach Adam-Scale speed when constrained to inference‑level memory.

\textbf{Batch‑based zeroth‑order optimizers}~A batch‑oriented refinement of ZO methods estimates gradients by evaluating a batch of forward step.  Recent work like ReLIZO~\citep{ReLIZO} reuses or correlates directions to boost sample efficiency. Augmented Random Search~\citep{ARS} and Evolution Strategies ~\citep{sun2022bbtv2,sun2022black} adopt similar population‑based updates and are popular in reinforcement learning. However, such approaches generally lack specialized optimizations for accelerating the batched forward process, leading to limited speed when fine-tuning large models.

\begin{algorithm}[t]
\begin{algorithmic}[1]
\Require parameters $\theta \in \mathbb{R}^d$, loss $L : \mathbb{R}^d \rightarrow \mathbb{R}$, step budget $T$, perturbation scale $\epsilon$, batch size $N$, learning rate schedule $\{\eta_t\}$

\For{$t=1,...,T$}
    \State $\ell$, $\theta$, $seeds$ $\leftarrow$ BatchPerturbParameters($\theta$, $\epsilon$, $N$)
    \State $std$ $\leftarrow$ standard deviation of $\ell$
    \State $projected\_grad \leftarrow (\ell - \mathcal{L}(\theta; \mathcal{B})) / (N * std)$
    \State BatchUpdateParameter($projected\_grad$, $seeds$, $\theta$, ${\eta_t}$)
\EndFor
\Function{BatchPerturbParameter}{$\theta$, $\epsilon$, $N$}
    \State Sample batch $ \mathcal{B} \subset \mathcal{D}$; obtain input $X$ and first layer weights $W^{(1)}$
    \State Initialize random seeds $seeds \leftarrow \{s_1, \dots, s_N\}$
    \State Generate perturbation vectors $u \in \{\pm 1\}^{N \times d}$ using $seeds$
    \State Compute unperturbed activations $F^{(1)} \leftarrow W^{(1)}X$
    \State Compute perturbed activations: $Y^{(1)}_i \leftarrow F^{(1)} + \epsilon(u_i \odot X),\, i=1,\dots,N$
    \State Concatenate activations $Y^{(1)} \leftarrow [Y^{(1)}_1;\dots;Y^{(1)}_N]$
    \For{$j=2,3,\dots$}
        \State $F^{(j)} \leftarrow W^{(j)}Y^{(j-1)}$ \Comment{Compute unperturbed activations at layer $j$}
        \State $P^{(j)} \leftarrow \epsilon(u \odot Y^{(j-1)})$ \Comment{Compute perturbations at layer $j$}
        \State $Y^{(j)} \leftarrow F^{(j)} + P^{(j)}$ \Comment{Compute perturbed activations in parallel at layer $j$}
    \EndFor
    \State $\ell \leftarrow \mathcal{L}(Y^{(final)}; \mathcal{B})$ \Comment{Compute final losses in parallel}
    \State \Return $\ell$, $\theta$, $seeds$
\EndFunction

\Function{BatchUpdateParameter}{$projected\_grad$, $seeds$, $\theta$, ${\eta}$}
    \For{$idx, s \in \texttt{enumerate}(seeds)$}
        \State Reset random number generator with seed $s$ 
        \Comment{For sampling $u$}
        \For{$\theta_i \in \theta$}
            \State $u \sim \text{Uniform}(\{+1, -1\})$
            \State $\theta_i \leftarrow \theta_i - \eta \ast  projected\_grad_{idx} \ast u$
        \EndFor
    \EndFor
\EndFunction

  \caption{\name}
  \label{alg:FZOO}
  \end{algorithmic}
\end{algorithm}

\section{Methods}\label{sec_methods}

\begin{figure*}[h]
\centering
\vskip -0.15in
\includegraphics[width=1\linewidth]{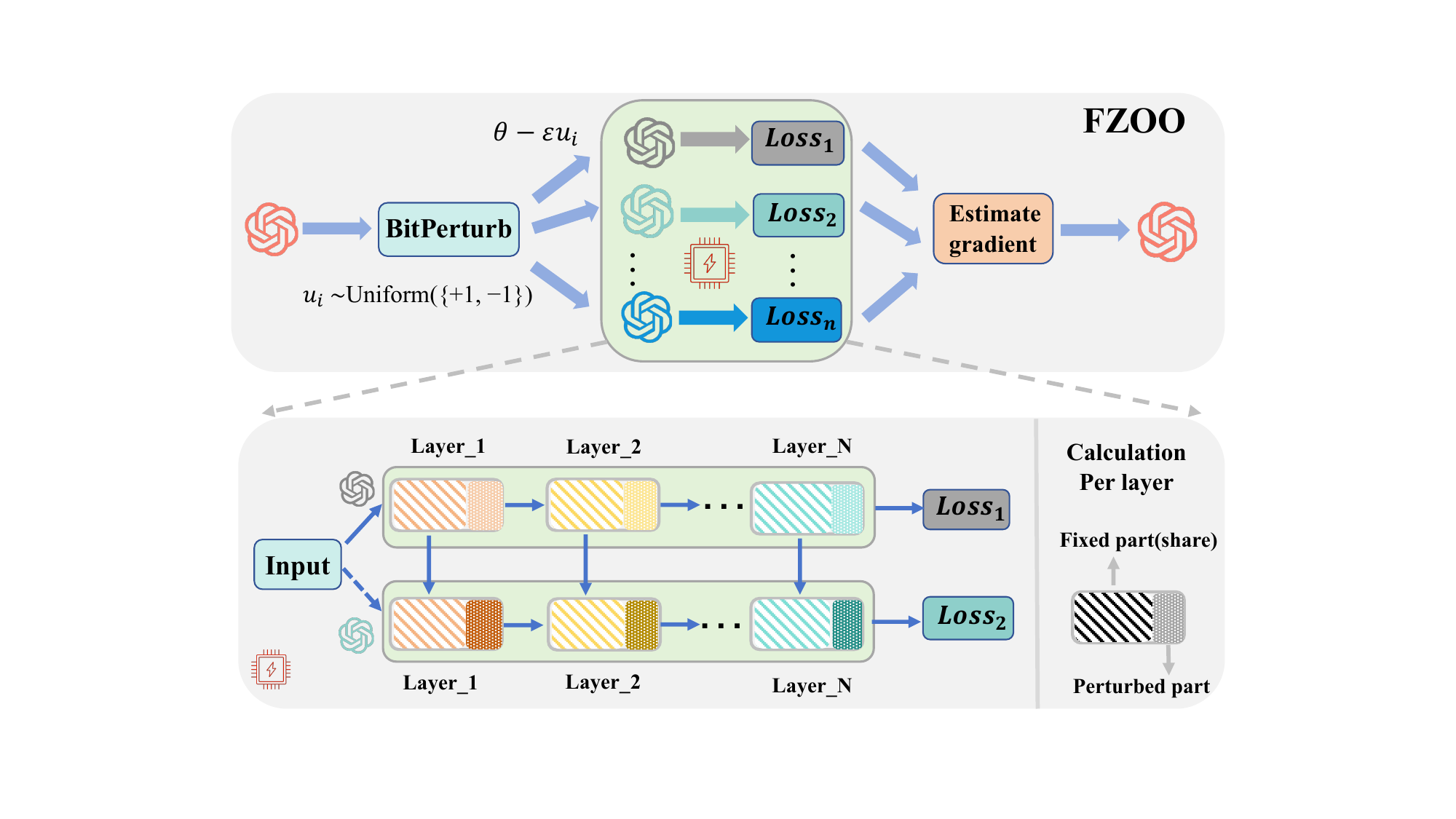}
\vskip -0.4in
\caption{Structure of the of FZOO. The bottom half depicts the toy example of the efficient implementation of batched forward passes.}
\label{fig:fzoo}
\vskip -0.1in
\end{figure*}

In the following, we first briefly review the classical zeroth‑order
gradient estimator employed in MeZO
(Section~\ref{sec_prelim}).  
We then present the complete workflow of our
\textbf{FZOO} optimizer (Section~\ref{sec_fzoo}), shown in Figure~\ref{fig:fzoo},  introduce a parallel perturbation strategy that accelerates computation
(Section~\ref{sec_parallel}),  
and finally offer a theoretical analysis of the proposed method
(Section~\ref{Equivalence2normalized-SGD},~\ref{sec_convergence}).

\subsection{Preliminaries}\label{sec_prelim}

Consider a labelled dataset
\(\mathcal{D}=\{(x_i,y_i)\}_{i=1}^{|\mathcal{D}|}\) and a mini‑batch
\(\mathcal{B}\subset\mathcal{D}\) of size \(B\).
Let \(\theta\in\mathbb{R}^{d}\) denote all trainable parameters of the
LLM, and let \(L(\theta;\mathcal{B})\) be the empirical loss on
\(\mathcal{B}\).

\begin{definition}[Classical ZO gradient estimation]\label{def:classical_zo}
Given a perturbation radius \(\epsilon>0\) and 
\(z\in\mathbb{R}^{d}\) sampled as \(z\sim\mathcal{N}(0,I_{d})\),
Classical ZO estimates the gradient on \(\mathcal{B}\) via
\begin{equation}\label{eq:classical_zo}
  \hat\nabla L(\theta;\mathcal{B})
  \;=\;
  \frac{L(\theta+\epsilon z;\mathcal{B})-
        L(\theta-\epsilon z;\mathcal{B})}{2\epsilon}\,
  z
  \;\;\approx\;\;
  zz^{\!\top}\,\nabla L(\theta;\mathcal{B}).
\end{equation}
Averaging \eqref{eq:classical_zo} over \(N\) i.i.d.\ draws
\(\{z_i\}_{i=1}^{n}\) yields the \(n\)-ZO estimator
  $\hat\nabla_{N}L=\frac1N\sum_{i=1}^{N}\hat\nabla_i L$.
\end{definition}

\textbf{Computational cost.} A single ZO estimate needs exactly \emph{two forward passes} and does not store activations for backpropagation. MeZO keeps memory usage at the inference level by passing random \emph{seeds} instead of full perturbation \(z\). MeZO further observes that choosing \(N>1\) improves statistical stability but increases computation by a factor of \(N\); therefore it
fixes \(N=1\).

\textbf{From Classical ZO to ZO‑SGD.} Replacing the back‑propagation gradient in SGD with the Classical ZO estimate gives the zeroth‑order update $\theta_{t+1}
  \;=\;
  \theta_t
  \;-\;
  \eta_t\,
  \hat\nabla L\bigl(\theta_t;\mathcal{B}_t\bigr)$.
MeZO realises this update in‑place with the memory tricks above and
serves as the baseline for our forthcoming improvements.

\subsection{Fast Zeroth‑Order Optimization}
\label{sec_fzoo}

MeZO employs a fixed learning rate.
Instead, FZOO dynamically tunes its step size using the standard deviation of loss values gathered from multiple forward passes within each mini-batch. 
For the enhanced computational efficiency, FZOO also incorporates a one-sided gradient estimation strategy and \textbf{we first utilize Rademacher random vectors for perturbations}. Letting \(u_1,\dots,u_N\) be $N$ i.i.d Rademacher random vectors in $\mathbb{R}^d$, we construct our gradient estimation by function value queries
\[
  l_i = L(\theta_t+\epsilon u_i;\mathcal{B}_t), 
  \quad \mbox{ and } \quad
  l_0 = L(\theta_t;\mathcal{B}_t).
\]
That is, the gradient estimate $g_t$ is then computed by averaging $N$ one-sided difference estimates 
\begin{equation}\label{eq:g_hat_t_def}
  {g_t} \;=\; \frac{1}{\epsilon N}\sum_{i=1}^{N}(l_i - l_0)u_i.
\end{equation}
The estimated variance $\sigma_t^2$ is computed as:
\begin{equation}\label{eq:sigma}
    \sigma_t^2 = \frac{1}{N-1} \sum_{i=1}^{N}\left(l_i - \frac{1}{N}\sum_{j=1}^{N} l_j\right)^2.
\end{equation}
Our FZOO updates the parameters according to~\eqref{eq:g_hat_t_def} and~\eqref{eq:sigma}:
\begin{equation}\label{eq:update}
\theta_{t+1} = \theta_t -  \eta_t\frac{ g_t}{\sigma_t},
\end{equation}
where $\eta_t$ is the step size.
The detailed implementation of \name is listed in Algorithm~\ref{alg:FZOO}.

The \eqref{eq:update} shows that FZOO  receives larger steps at flat regions (where $\sigma_t$ is small) and smaller steps at steep regions, mirroring Adam‑style adaptivity while keeping memory at the inference level.

\subsubsection{Movtivation of FZOO}

Adaptive first-order methods often estimate local curvature to scale updates. Similar adaptivity can be achieved by methods like normalized-SGD~\citep{bernstein2018signsgd}, which adjusts step sizes by normalizing the gradient, making it more memory-efficient compared to Adam. 
The parameter update follows normalized-SGD:
\begin{equation}\label{eq:normalized-SGD}
\theta_{t+1} = \theta_t - \eta_t \frac{{g}_t}{||{g}_t||},
\end{equation}
FZOO is inspired by  normalized-SGD. Proposition~\ref{prop:nsgd} shows that $ \sigma_t^2 = |g_t|^2 \cdot \epsilon^2 \cdot \frac{N-1}{N} $ which implies that FZOO is an extension of normalized-SGD to the ZO domain.

\textbf{Variants.}
\emph{FZOO‑R} re‑uses half of the losses from the previous mini‑batch:
\(
\sigma_t=\operatorname{Std}\bigl(\{l_i^{\text{curr}}\}\cup
\{l_i^{\text{prev}}\}\bigr),
\)
achieving a full‑batch variance estimate with only half the forward
evaluations.  

\subsection{Efficient Implementation of Batched Forward Pass}
\label{sec_parallel}

The original ZO pipeline applies a perturbation and then runs a separate forward pass, which cannot take advantage of CUDA parallel computing.  For layer $j$ we split the computation into an unperturbed part $F^{(j)}$ and a perturbation part $P^{(j)}$.  Using Gaussian noise forces two full matrix–vector products, so batching recovers little speed‑up.  Replacing the noise with a Rademacher vector $u_i\in\{\pm1\}^d$ changes only the sign bits: $P^{(j)}$ is produced by a bit‑level sign flip that degenerates into a single add / subtract, so the kernel issues additions instead of a second matrix–vector multiply, making it markedly faster because addition is cheaper than multiplication on CUDA cores~\citep{williams2009roofline}.

Let \(X\) be the input and let \(W^{(j)}\) denote the weight matrix of layer \(j\). For the \textbf{first layer} we compute 
\[
  F^{(1)} = W^{(1)}X,\qquad
  Y^{(1)}_i = F^{(1)} + \epsilon\,(u_i\odot X),\; i=1,\dots,N,
\]
and we stack the perturbed outputs along the batch dimension:
\(Y^{(1)} = [Y^{(1)}_1;\dots;Y^{(1)}_N]\).

Let \(Y^{(j-1)}\) be the concatenated activations from the previous layer and
\(U=\mathrm{diag}(u_1,\dots,u_N)\) the block‑diagonal sign matrix that
broadcasts the \(u_i\) across the batch axis. For \textbf{subsequent layers} (\(j\ge2\)) we compute
\[
  F^{(j)} = W^{(j)}Y^{(j-1)},\qquad
  P^{(j)} = \epsilon\,(U\odot Y^{(j-1)}),\qquad
  Y^{(j)} = F^{(j)} + P^{(j)}.
\]

Because every layer’s \(N\) matrix multiplications are fused into a single CUDA kernel, wall‑time drops by a factor \(p\) relative to sequential evaluation.  
Combining kernel fusion (\(p\)), Rademacher vector perturbations (\(r\)), and the one‑sided estimator’s halved forward count (\(f\)) yields the overall speed‑up
\[
  \boxed{\,f \times \min(s,r)\,}.
\]

On OPT‑125M with $N=8$, our batched scheme($\min(s,r)$) delivers a $1.92\times$ speed‑up over the “8 perturbations + 8 forward passes” baseline.  Because the code still runs on vanilla \texttt{transformers}, even larger gains should be possible with high‑throughput runtimes such as vLLM (section~\ref{sec:memory_time}).
.

\subsection{Equivalence to normalized-SGD}
\label{Equivalence2normalized-SGD}


\begin{proposition}\label{prop:nsgd}
 Let the stochastic gradient estimation $g_t$ defined in \eqref{eq:g_hat_t_def} and the variance $\sigma_t$ defined in \eqref{eq:sigma}.
 Then it holds that
 \begin{align}
 \mathbb{E}\left[\norm{g_t}^2\right] =& \frac{N+d-1}{N} \norm{\nabla L(\theta_t , \mathcal{B}_t)}^2 + \gamma_t \quad \mbox{and} \quad \gamma_t = O(\epsilon) \label{eq:gt_norm}\\
     \mathbb{E}\left[\sigma_t^2\right] =& \epsilon^2 \cdot \norm{\nabla L(\theta_t, \mathcal{B}_t)}^2 + \zeta_t \quad \mbox{with} \quad \zeta_t = O(\epsilon^3). \label{eq:sig}
 \end{align}
\end{proposition}

\begin{remark}
Comparing \eqref{eq:sig} and \eqref{eq:gt_norm}, and ignoring the perturbation terms $\zeta_t$ and $\gamma_t$ (which are higher-order terms with respect to $\epsilon$), we can obtain that:
$$ \mathbb{E}\left[\norm{g_t}^2\right] = \frac{N+d-1}{N} \cdot \epsilon^{-2} \cdot \mathbb{E}\left[\sigma_t^2\right] $$
Since $\frac{N+d-1}{N} \cdot \epsilon^{-2}$ is a constant independent of the iteration number, $\frac{g_t}{\sigma_t}$ can be regarded as a kind of normalized stochastic gradient, scaled by a constant factor. This theoretically establishes the connection between FZOO and normalized-SGD.
\end{remark}

\subsection{Convergence Analysis}
\label{sec_convergence}

\begin{assumption}
Suppose that the loss function $L(\theta)$ is $\cL$-smooth, that is, for any $\theta_1,\theta_2\in\RR^d$, it holds that $L(\theta_2) \leq L(\theta_1) + \dotprod{\nabla L(\theta_1), \theta_2 - \theta_1} + \frac{\cL}{2}\norm{\theta_2 - \theta_1}^2$.
\end{assumption}
\begin{assumption}[Bounded Variance]\label{ass:var}
The stochastic gradient $\nabla L(\theta, \mathcal{B})$ has bounded variance:
$$ \mathbb{E}\left[\norm{\nabla L(\theta, \mathcal{B})}^2\right] \leq \norm{\nabla L(\theta)}^2 + \mathcal{V}^2, $$
where $\mathcal{V}^2$ is a constant. The above is a standard assumption for stochastic gradient descent.
\end{assumption}

\begin{theorem}[Convergence of FZOO]\label{thm:fzoo_convergence}
Let the objective function $L(\theta)$ be $\mathcal{L}$-smooth, and Assumption~\ref{ass:var} hold. The update rule is $\theta_{t+1} = \theta_t - \eta_t g_t$, where the effective step size $\eta_t = \frac{\eta}{\sigma_t}$ satisfies $\eta_t \leq \frac{N}{16d\mathcal{L}}$, and the learning rate factor $\eta$ is set as 
$ \eta = \left(\frac{L(\theta_1) - L(\theta^*)}{4d\mathcal{L} \mathcal{V}^2 \sum_{k=1}^{T} \sigma_k^{-2}}\right)^{1/2} $.
Then, after $T$ iterations, the FZOO algorithm satisfies:
\begin{align*}
\frac{1}{T}\sum_{t=1}^{T} \mathbb{E}\left[ \norm{\nabla L(\theta_t)}^2 \right] \leq& \frac{64\sigma_* \left(d\mathcal{L}(L(\theta_1) - L(\theta^*)) \sum_{t=1}^{T} \sigma_t^{-2} \right)^{1/2}}{T} \\
&+ \frac{4\sigma_* \epsilon^2}{T} \left(\frac{d^3\mathcal{L}^3}{8N} + \frac{Nd\mathcal{L}^3}{2}\right) \left(\frac{\Big(L(\theta_1) - L(\theta^*)\Big) \sum_{t=1}^{T} \sigma_t^{-2}}{4d\mathcal{L} \mathcal{V}^2 }\right)^{1/2}
\end{align*}
where $\theta^*$ denotes the optimal solution (minimum) of $L(\theta)$, and $\sigma_* = \max_{t \in \{1, \dots, T\}} \sigma_t$.
\end{theorem}
This theorem shows that the FZOO algorithm converges to a stationary point whose gradient norm close to zero.


\section{Experiments}
\label{sec_experiments}

Large language models (LLMs) fall into two architectural families. Encoder–decoder models—BERT~\citep{BERT}, ALBERT~\citep{ALBERT}, etc.—tackle language understanding and are trained with a masked‑token objective (Section~\ref{ssec:masked lauguage}). Decoder‑only models—GPT~\citep{gpt2,GPT3}, OPT~\citep{opt}, LLaMA~\citep{Touvron2023LLaMAOA}, Phi~\citep{li2023textbooks,gunasekar2023textbooks}, and others—focus on text generation and are trained autoregressively (Section~\ref{ssec:auto-regressive}). We benchmark FZOO under full‑parameter fine‑tuning on both model families, covering differentiable and non‑differentiable tasks (Section~\ref{section_F1}), report its memory and wall‑clock speed (Section~\ref{sec:memory_time}), and compare it with standard ZO baselines (Section~\ref{ssec:ZO_variants}). Section~\ref{orthogonality_hyperparameter} further shows that FZOO is complementary to parameter‑efficient methods such as prefix‑tuning~\citep{li-liang-2021-prefix}.


\begin{figure*}[ht]
\centering
\includegraphics[width=1\linewidth]{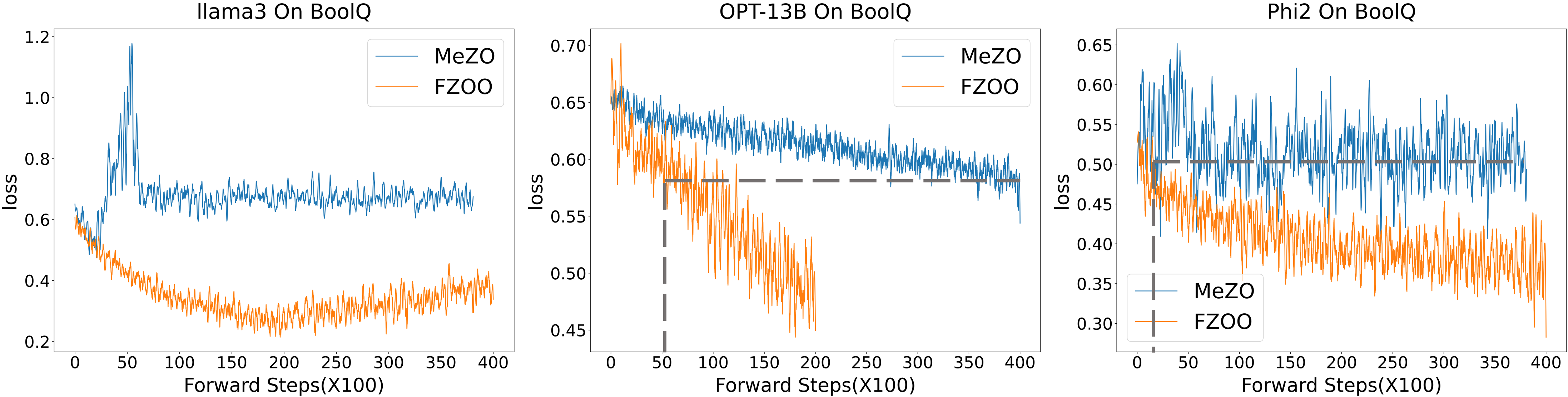}
\vspace{-6mm}
\caption{Training loss curves when using MeZO and FZOO to fine-tune different LLMs on the BoolQ dataset. 
}
\label{fig:roberta-16}
\vskip -0.05in
\end{figure*}

\subsection{Masked Language Models}
\label{ssec:masked lauguage}
We perform experiments on the RoBERTa-large 350M model~\citep{roberta} across three NLP task paradigms: sentence classification, multiple choice and text generation. 
 Following the experimental setup outlined in~\citep{mezo}, we investigate both few-shot and many-shot settings, sampling $k$ examples per class with $k = 16$ (results presented in Table~\ref{tab:roberta_k16}) and $k = 512$ (results provided in Appendix~\ref{appendix_robert}).

According to existing analysis, the cost of a single backpropagation is about three times that of a forward propagation~\citep{alman2024fine}, so the time required for each parameter update of Adam is equivalent to four forward propagations. Based on this, we plotted the loss curve in Figure \ref{fig:fzoo_adam}. As can be seen from Figure \ref{fig:fzoo_adam}, FZOO can reduce loss faster than MeZO, with an acceleration ratio of 18 times (with 1.92x parallel acceleration, it can potentially outperform Adam in wall-clock time). And its loss convergence speed is close to Adam, fully demonstrating FZOO's excellent performance in training efficiency.
\begin{table*}[t]
\vspace*{-2mm}
\centering
\caption{Experiments on RoBERTa-large (350M parameters, k=16). All reported numbers are averaged accuracy across 5 runs. In notations like (2xM), M represents the memory overhead required for inference.}
\vspace*{2mm}
\scalebox{0.85}{
    \begin{tabular}{lccccccc}
    \toprule
    Task Type & \multicolumn{1}{c}{\textbf{SST-2}} & \multicolumn{1}{c}{\textbf{SST-5}} & \multicolumn{1}{c}{\textbf{SNLI}}  & \multicolumn{1}{c}{\textbf{MNLI}} & \multicolumn{1}{c}{\textbf{RTE}} & \multicolumn{1}{c}{\textbf{TREC}} & \multicolumn{1}{c}{\textbf{Average}}\\
    & \multicolumn{2}{c}{------ sentiment ------} & \multicolumn{3}{c}{------ natural language inference ------} & \multicolumn{1}{c}{--- topic ---}\\
    \midrule
    Zero-shot & 79.0 & 35.5 & 50.2 & 48.8 & 51.4 & 32.0 & 49.5\\
    LP        & 76.0 & 40.3 & 66.0 & 56.5 & 59.4 & 51.3 & 58.3\\
    \midrule
    HiZOO~(2xM)   & 93.2 & 46.2 & 74.6 & 64.9 & 66.8 & 79.8 & 70.9\\
    ZO-Adam~(3xM) & 90.3  & 38.8  & 69.1  & 59.2  & 61.4  & 52.4  & 61.7 \\
    FT~(6xM)      & 91.9 & 47.5 & 77.5 & 70.0 & 66.4 & 85.0 & 74.9\\
    \midrule
    MeZO(prefix)   & 90.8 & 45.8 & \textbf{71.6} & 63.4 & \textbf{65.4} & 80.3 & 69.6\\
    HiZOO(prefix) & 92.3  & \textbf{47.2}  & 68.8  & 61.6  & \textbf{65.4}  & 82.0  & 69.6 \\
    FZOO(prefix)      & \textbf{92.9} & 43.7 & 70.8 & \textbf{67.0} & 64.6 & \textbf{82.8} & \textbf{70.3}\\
    \midrule
    MeZO      & 90.5 & 45.5 & 68.5 & 58.7 & 64.0 & 76.9 & 67.4\\
    FZOO      & \textbf{93.3} & \textbf{47.6 }& \textbf{75.9} & \textbf{64.9} & \textbf{67.9} & \textbf{78.8} & \textbf{71.4}\\
    \bottomrule
    \end{tabular}}
    \label{tab:roberta_k16}
\end{table*}

Table~\ref{tab:roberta_k16} shows that when optimizing the RoBERTa-large model with MeZO and \nameo, the performance of \nameo\ is on average \textbf{5.9\%} higher than that of MeZO on multiple datasets across different NLP tasks. Specifically, \nameo\ outperforms MeZO more than \textbf{10.7\%} in both the SNLI and MNLI dataset, showing comparable performance to HiZOO.

\begin{table*}[h]
\centering
\vspace{-5pt} 
\caption{Experiments on three different models (with $1000$ examples) : Classification (SST-2, RTE, CB, BoolQ, WSC, WIC, MultiRC); Multiple Choice (COPA, ReCoRD); Generation (SQuAD, DROP). We highlight the best results between MeZO, HiZOO-L and FZOO in bold to facilitate comparison. \label{tab:Llama3_phi2}}
\vspace{1.5pt} 

\resizebox{\textwidth}{!}{%
  \setlength{\tabcolsep}{4pt}

  \begin{tabular}{lc*{14}{c}} 
    \toprule

    & \textbf{Model} & \textbf{Method} & \textbf{SST-2} & \textbf{RTE} & \textbf{CB} & \textbf{BoolQ} & \textbf{WSC} & \textbf{WIC} & \textbf{MultiRC} & \textbf{COPA} & \textbf{ReCoRD} & \textbf{SQuAD} & \textbf{DROP} & \textbf{Average}\\
    \midrule

    &Phi-2 &MeZO    &86.6 &67.1 &75.0 &72.4 &59.6 &54.4 &78.2 &\textbf{86.0} &71.7 &85.7 &\textbf{37.8} &70.7\\
    &Phi-2 &HiZOO-L &\textbf{88.9} &68.9 &75.2 &72.0 &62.4 &59.2 &79.2 &\textbf{86.0} &\textbf{72.1} &85.7 &36.2 &71.4 \\
    &Phi-2 &FZOO   &87.4 &\textbf{70.4} &\textbf{83.9} &\textbf{79.3} &\textbf{61.5} &\textbf{56.7} &\textbf{81.3} &\textbf{86.0} &72.0 &\textbf{86.7} & 37.4 &\textbf{73.0}\\ 
    \midrule
    
    &Llama3 &MeZO   &92.2 &74.4 &\textbf{69.6} &76.7 &63.5 &57.8 &77.6 &88.0 &\textbf{85.6} &86.7 &\textbf{57.1} &75.4\\
    &Llama3 &HiZOO-L&\textbf{94.3} &75.1 &\textbf{69.6} &77.1 &63.5 &57.7 &77.9 &\textbf{89.0} &\textbf{85.6} &87.7 &49.4 &75.2 \\
    &Llama3 &FZOO  &\textbf{94.3} &\textbf{77.6} &\textbf{69.6} &\textbf{81.8} &\textbf{65.4} &\textbf{60.8} &\textbf{81.5} &88.0 &85.3 &\textbf{87.9} &56.5 &\textbf{77.2}\\ 
    \midrule

    &OPT-13B &MeZO    &91.4 &66.1 &66.0 &67.6 &63.5 &59.4 &57.3 &88.0 &\textbf{81.7} &84.7 &\textbf{30.9} &68.8 \\
    &OPT-13B &HiZOO-L &92.1 &68.2 &67.9 &66.6 &\textbf{65.4} &59.4 &61.1 &\textbf{89.0} &81.1 &63.6 &22.7 &67.0\\
    &OPT-13B &FZOO   &\textbf{93.7} &\textbf{71.1} &\textbf{69.6} &\textbf{72.2} &63.5 &\textbf{60.5 }&\textbf{66.0} &87.0 &81.0 &\textbf{84.8} &28.7 &\textbf{70.7}\\

    \bottomrule
  \end{tabular}%
} 
\vspace{-5pt}
\end{table*}

\subsection{Auto-Regressive Language Models}
\label{ssec:auto-regressive}
Next, we expand the scope of our experiments by testing the Phi-2(2.7B), Llama3(8B) and OPT(13B) models on the same NLP tasks.
The results of the experiment in Table~\ref{tab:Llama3_phi2} show that FZOO outperforms MeZO in most cases, with an average accuracy increase of \textbf{2.75\%}. Aside from the ReCoRD and DROP datasets, FZOO surpasses MeZO in nearly all other cases.

\textbf{\nameo\ significantly accelerates convergence speed during full-parameter tuning.} As shown in Figure \ref{fig:roberta-16}, \nameo\ not only achieves \textbf{8}\boldsymbol{$\times$} speedup over MeZO on average in the context of full parameter tuning, but also has lower training loss. In addition, FZOO ultimately achieved a higher absolute accuracy improvement of \textbf{7.4\%} than MeZO on the BoolQ task. 

\textbf{\name is able to adapt to large models with up to 66B parameters and maintains excellent performance.} As shown in Table~\ref{tab_opt66}, on the OPT-66B model, \name outperforms MeZO with up to \textbf{13.2\%} increase and  \textbf{2.43\%} increase on average.

\subsection{Training with Non-Differentiable Objectives}
\label{section_F1}

Our proposed \name  estimates the gradient by performing multiple forward propagation processes, so the objective function can be a non-differentiable function. Based on the MeZO~\citep{mezo} setup, we conduct some experiments using F1 as the optimization objective. As shown in Table~\ref{tab_F1}, FZOO performs consistently well across OPT models of different scales, surpassing both MeZO and HiZOO-L, and achieving an average F1 improvement of \textbf{5.53\%} over MeZO.

\begin{table}[htbp]
  \centering
  \vspace{-10pt}
  \begin{minipage}[t]{0.48\linewidth} 

\centering
\caption{
    Experiments on OPT-30B  and OPT-66B (both use FT). 
}
\label{tab_opt66}
\vspace{1.8pt}
\resizebox{1.0\textwidth}{!}{ 
    \begin{tabular}{lcccccccc}
    \toprule
     Task  & \textbf{SST-2} & \textbf{RTE}  & \textbf{WSC} & \textbf{WIC}  & \textbf{Average} \\
    \midrule
    30B MeZO & 90.6 & 66.4  & \textbf{63.5} &56.3 & 69.2 \\
    30B HiZOO-L &90.3 & 69.3  & \textbf{63.5} &53.4 &69.3\\
    30B FZOO &\textbf{91.2} & \textbf{69.3} & \textbf{63.5} &\textbf{60.2} &\textbf{71.1}\\

    \midrule
    66B MeZO & 91.2 & 65.7  & \textbf{63.5} &58.9 & 69.8 \\
    66B HiZOO-L &88.9 &66.4  &59.6 &58.6 &68.4\\
    66B FZOO &\textbf{93.6} & \textbf{74.4}  & 58.6 &\textbf{59.2} &\textbf{71.5}\\

    \bottomrule
    \end{tabular}
}
  \end{minipage}
  \hfill
  \begin{minipage}[t]{0.48\linewidth} 
    \centering

    \caption{Non-differentiable optimization objectives (F1): Performance comparison on the SQuAD Task.}
\label{tab_F1} 
\vspace{1.8pt} 
\resizebox{1.0\textwidth}{!}{ 
    \begin{tabular}{lcccccc} 
    \toprule
    Model & \multicolumn{6}{c}{\textbf{OPT}} \\ 
    \cmidrule(lr){2-7}
    Size  & \textbf{125M} & \textbf{1.3B} & \textbf{2.7B} & \textbf{6.7B} & \textbf{13B} & \textbf{Average} \\ 
    \midrule
    Zero-shot & 9.7  & 27.2 & 29.9 & 36.6 & 46.2 & 29.9 \\ 
    MeZO      & 44.1 & 72.2 & 77.6 & \textbf{80.1} & 78.5 & 70.5 \\ 
    HiZOO-L   & 37.6 & 71.9 & 77.5 & 79.5 & 81.8 & 69.7 \\
    FZOO      & \textbf{51.0} & \textbf{74.6} & \textbf{81.8} & 79.8 & \textbf{84.7} & \textbf{74.4} \\ 
    \bottomrule
    \end{tabular}
}
  \end{minipage}
  \vspace{-10pt}
  
\end{table}


\subsection{Memory Usage and Time Efficiency Analysis}
\label{sec:memory_time}

\begin{figure*}[ht]
\centering
\includegraphics[width=0.7\linewidth]{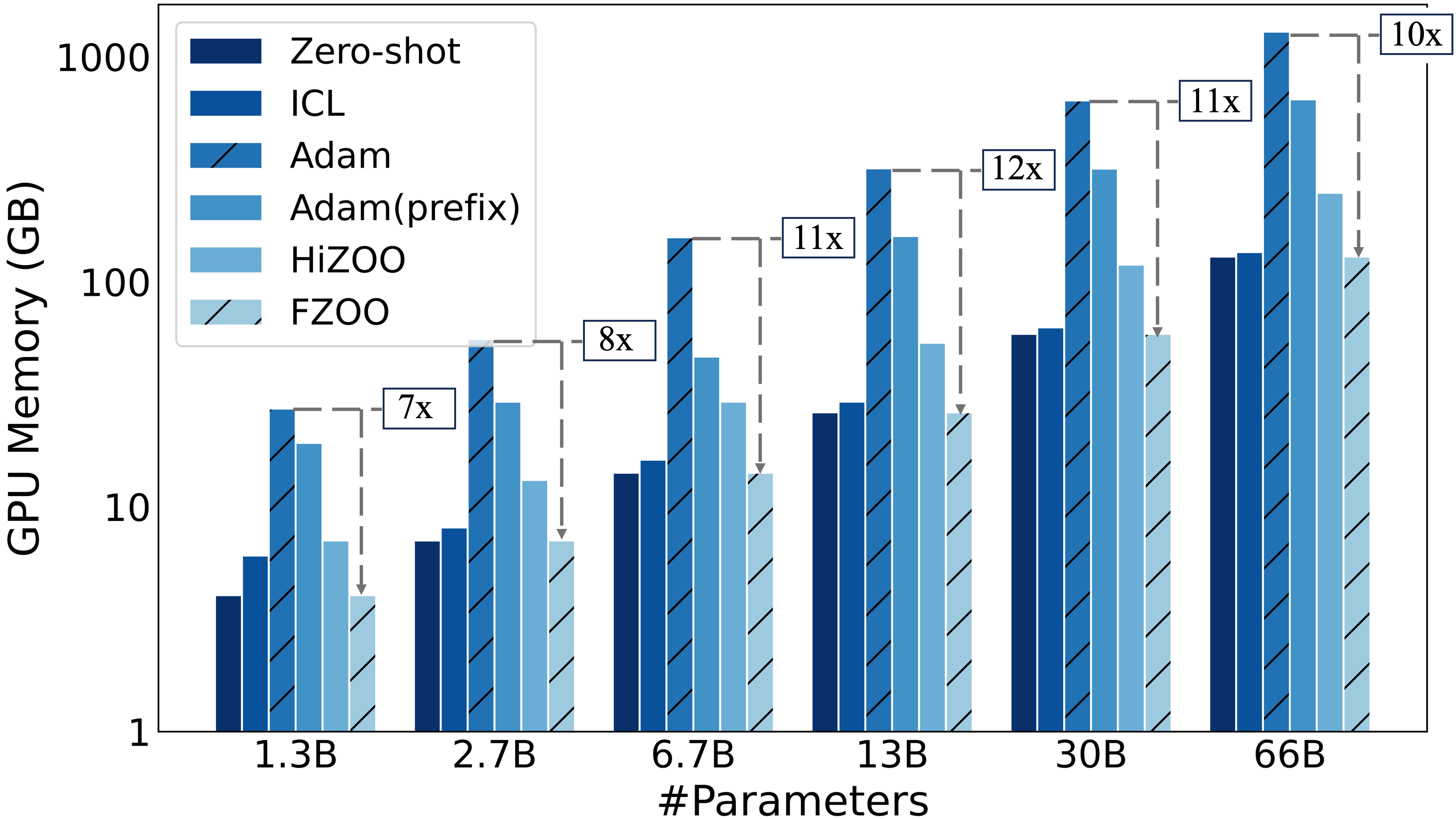}
\caption{GPU memory consumption with different OPT models and tuning methods on MultiRC (400 tokens per example on average). More details can be found in Appendix~\ref{app:memory_time}.}
\label{fig:memory}
\vskip -0.05in
\end{figure*}

\textbf{Memory Usage}~~~~As shown in Figure~\ref{fig:memory}, FZOO and MeZO have the same inference-level memory and do not need to store additional content like HiZOO. Even if Adam is trained with PEFT like prefix, it still requires several times more memory than FZOO.

\begin{table*}[htbp]
  \centering
  \vspace{-3pt}
  \caption{Wallclock time per step between Adam, MeZO and FZOO(N=8). See Appendix~\ref{app:train_time} for more. }
  \vspace{4pt}
  \scalebox{1}{
    \begin{tabular}{lccc} 
    \toprule
    \textbf{Method} & \textbf{OPT-125M} & \textbf{RoBERTa-large} & \textbf{OPT-1.3B} \\
    \midrule
    Adam & 0.1982s & 0.3930s & 0.5814s \\
    MeZO & 0.1368s & 0.4305s & 0.7218s \\
    FZOO & 0.3835s & 0.6052s & 1.6628s \\
    \bottomrule
    \end{tabular}
  }
  
  \label{tab:app_train_time}
\end{table*}

\begin{table*}[htbp]
  \centering
  \caption{Actual and potential speedup of FZOO for different tasks. Adam commonly achieves 20x speedup than MeZO.}
  \vspace{4pt}
  \begin{tabular}{lcccc}
    \toprule
    \textbf{Task(model)} & \textbf{SNLI (RoBERTa-large)} & \textbf{COPA (Phi-2)} & \textbf{WIC (OPT-13B)} & \textbf{CB (Llama3)} \\
    \midrule
    FZOO        & 20× & 10× & 9×  & 8×  \\
    Potential   & 40× & 20× & 18× & 16× \\
    \bottomrule
  \end{tabular}
  \label{tab:task_speedups}
  \vspace{-4mm}
\end{table*}

\textbf{Time Efficiency}~~~~Adaptive step‑size scheduling delivers up to $20\times$ speed‑up in total steps over MeZO (Table~\ref{tab:task_speedups}). One FZOO step bundles 9 forward passes—$4.5\times$ more than MeZO—yet consumes only $3\times$ the wall‑clock time (Appendix~\ref{app:train_time}), highlighting the gain from our parallel implementation. 

\subsection{Comparison with other ZO variants}
\label{ssec:ZO_variants}

We further compare \nameo\ with a wider range of ZO optimization methods~\citep{zobenchmark}. As shown in Table~\ref{benchmark}, \nameo\ consistently outperforms all baselines. Compared to ZO-Adam, our \nameo\ achieves an average accuracy improvement of \textbf{5.77\%} under full-parameter tuning, and \textbf{4.21\%} under prefix-tuning, while using \textbf{40.5\%} of the GPU memory.

\begin{table*}[h]
\centering
\vspace{-15pt}
\caption{Performance comparison on SST2(Robert-Large and OPT-1.3B) and COPA(OPT-13B) using different ZO methods. Memory and runtime cost are multiples of
ZO-SGD.}
\vspace{7pt}
\scalebox{0.88}{
\begin{tabular}{lcccccccccc}
\toprule
\textbf{Model/Task} & \multicolumn{2}{c}{\textbf{Roberts-Large}} & \multicolumn{2}{c}{\textbf{OPT-1.3B}} & \multicolumn{2}{c}{\textbf{OPT-13B}} & \textbf{Average} &\textbf{Memory} & \textbf{Runtime}\\
\cmidrule(lr){2-3} \cmidrule(lr){4-5} \cmidrule(lr){6-7}
 & FT & prefix & FT & prefix & FT  & prefix &  \\
\midrule

ZO-SGD & 89.4  & 90.0 & \textbf{90.8}  & 91.4 & \textbf{90.0}  & 79.0 & 88.4 & 1.0x & 1.0x\\
ZO-SGD-MMT & 89.6 & 89.1 & 85.2  & 91.2 & 87.0  & 85.0 & 87.8 & 1.56x & 1.0x\\
ZO-SGD-Cons & 89.6  & 89.1 & 88.3  & 88.1 & 82.0  & 84.0 & 86.8 & 1.0x & 2.49x\\
ZO-SGD-Sign & 52.5  & 53.6 & 87.2  & 89.5 & 80.0  & 78.0 & 73.4 & 1.0x & 1.0x \\
ZO-Adam & 89.8  & 90.2 & 84.4  & 91.4 & 82.0 &  79.0 & 86.1 & 2.47x & 1.04x \\
HiZOO & 93.2  & 92.7 & 90.7  & 91.4 & 88.0  & \textbf{87.0} & \textbf{90.5} & 2.04x & 1.37x\\
HiZOO-L & 92.5  & 92.7 & 90.7  & 91.4 & 88.0  & \textbf{87.0} & 90.4 & 1.12x & 1.39x \\
FZOO & \textbf{93.3}  & \textbf{92.9} & 90.7  & \textbf{91.7} & 87.0  & \textbf{87.0} & 90.4 & 1.0x & 0.56x \\

\bottomrule
\end{tabular}
}
\vspace{-7pt}
\label{benchmark}
\end{table*}

\subsection{Orthogonality and Hyperparameter Analysis}
\label{orthogonality_hyperparameter}

\vspace{-7pt}
\begin{figure}[h]
    \hspace{0.1in}
    \begin{minipage}[t]{0.46\textwidth}
      \centering
      \includegraphics[width=0.9\linewidth]{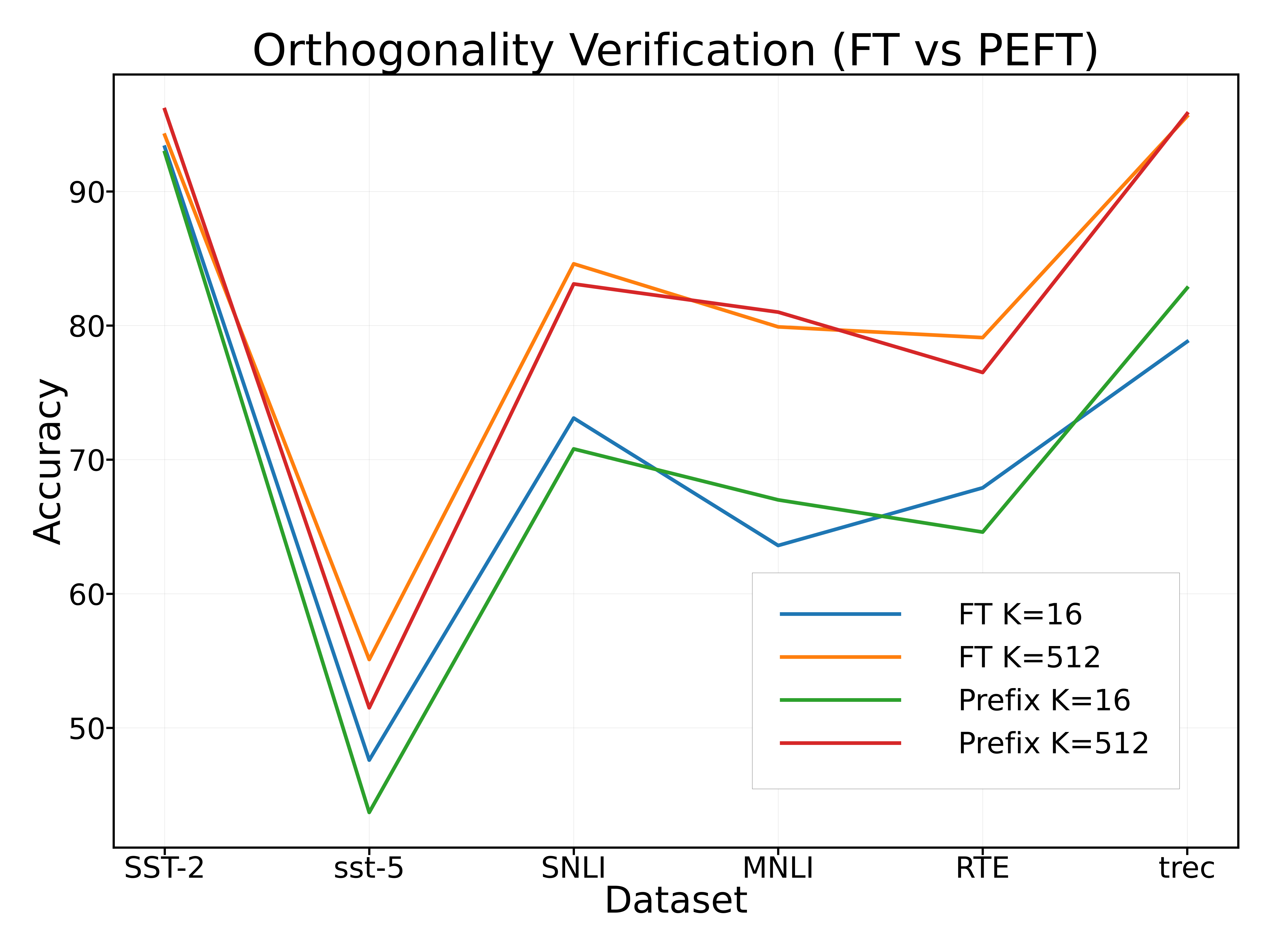}
      \vspace{-5pt}
      \caption{FZOO FT vs PEFT (prefix) on Roberta-large. We use \name(prefix) to fine-tune Roberta-large on multiple datasets. More results can be found in Appendix \ref{appendix_robert}. 
      }
      \vspace{-3pt}
      \label{fig:smooth_scale}
    \end{minipage}
    \hspace{0.1in}
    \begin{minipage}[t]{0.46\textwidth}
      \centering
      \includegraphics[width=0.9\linewidth]{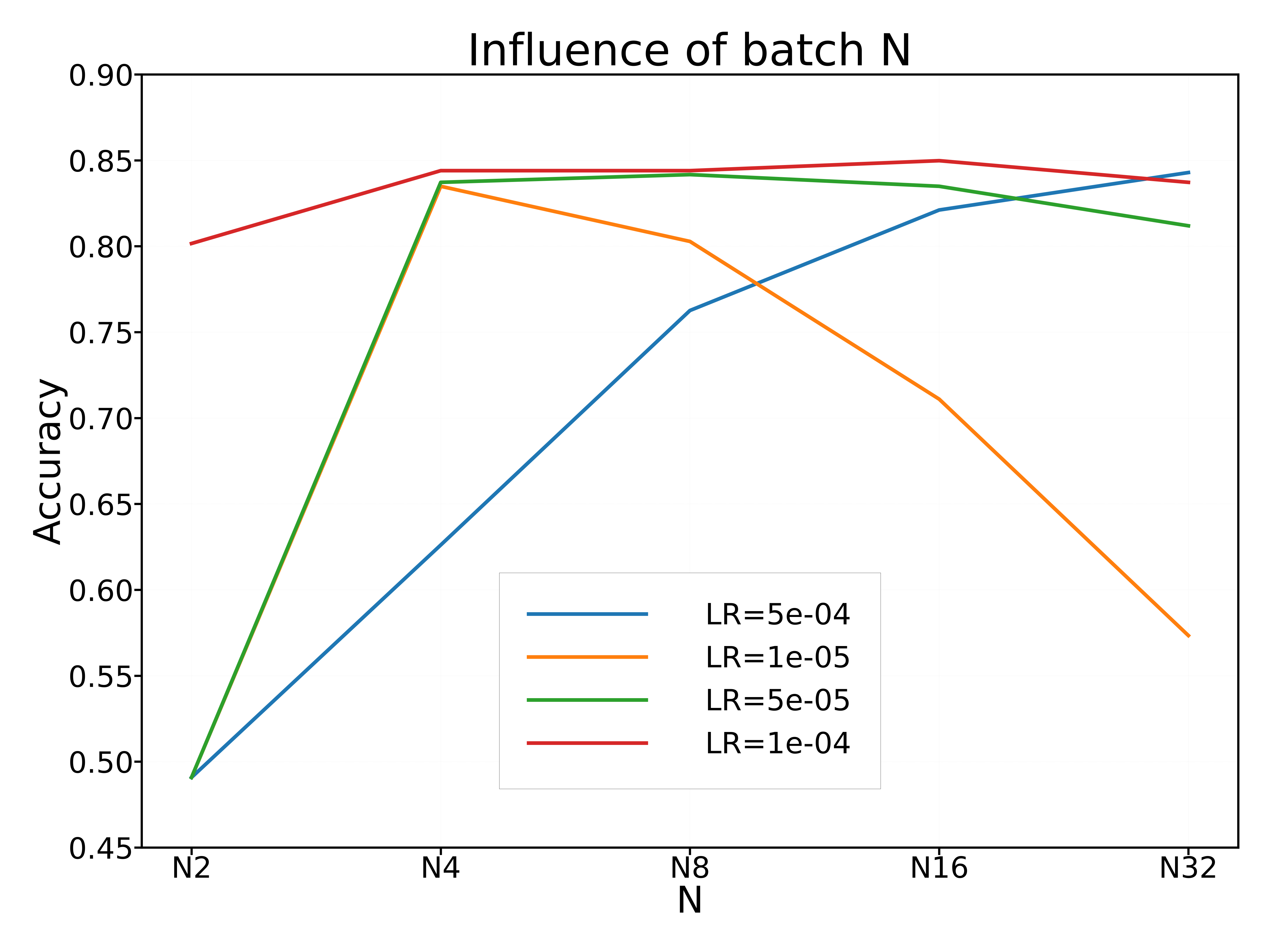}
      \vspace{-5pt}
      \caption{Accuracy Comparison under Different Perturbation Batch Sizes N for FT. The relevant experimental results are detailed in Appendix~\ref{app:smooth_scale}. }
      \vspace{-4pt}
      \label{fig:estimation_times}
    \end{minipage}
    
\end{figure}
\vspace{-2pt}

\textbf{Orthogonality to Fine-Tuning Strategies}~~~~Our method is an optimization algorithm that focuses on how to update parameters, and is orthogonal to fine-tuning strategies that decide which parameters to update. We validate its compatibility by FT or Prefix on the roberta-large model.

\textbf{Influence of the Number of Perturbations $N$ per Step}~~~~To further explore the trade-off between computational efficiency and convergence performance of the optimizer, we conducted a systematic evaluation on OPT-125M using the SST-2 dataset. As shown in Figure~\ref{fig:estimation_times}, increasing the number of perturbations $N$ per step can improve accuracy under certain hyperparameter settings, but also raises per-step computational cost.

\section{Conclusion}
\label{sec_conclusion}

We introduce FZOO, a novel zeroth-order optimizer achieving Adam-scale convergence speed for fine-tuning LLMs with inference-level memory. FZOO employs batched one-sided estimates and accelerated per-batch computations to significantly reduce forward passes. We provide theoretical analysis for its update rule and convergence. Experiments demonstrate FZOO converges in far fewer steps than MeZO, yielding superior performance across diverse LLMs. FZOO also seamlessly integrates with PEFT, enhancing practical fine-tuning.

\bibliographystyle{plain}
\bibliography{paper_ref}

\appendix
\clearpage
\input{appendix}

\end{document}

%% file: math_commands.tex
\usepackage{amsmath,amsfonts,bm}

\def\eqref#1{equation~\ref{#1}}

\def\1{\bm{1}}

\DeclareMathAlphabet{\mathsfit}{\encodingdefault}{\sfdefault}{m}{sl}
\SetMathAlphabet{\mathsfit}{bold}{\encodingdefault}{\sfdefault}{bx}{n}



%% file: appendix.tex
\section{Related Works}
\label{app_related_works}

\subsection{First-order Optimizer Used in LLMs}

In the development of deep learning, various optimization algorithms have been proposed to improve the efficiency and effectiveness of model training. Early methods include gradient descent (GD) and its improved versions, momentum method, Adagrad~\citep{Adagrad} and ADADELTA~\citep{zeiler2012adadelta}. These methods have their own characteristics, but they have played an indispensable role in promoting technological progress in key fields such as computer vision. As the scale of models rapidly expands, the number of parameters surges, and the structure becomes increasingly complex, the applicability of traditional optimization methods in large model training has been severely tested. In this context, Adam has gradually become the mainstream choice for training and fine-tuning large-scale models due to its efficient convergence performance. To further improve the generalization ability, AdamW \citep{adamw} was proposed.
These optimizers have an implicit upper limit on batch size, and this bottleneck is particularly problematic in training large models. To address this phenomenon, LAMB \citep{LAMB} was proposed, which uses a principled layer-wise adaption strategy to accelerate the training of large-scale models under substantial batch sizes.

\subsection{Zeroth-Order Optimizer}

Zero-order optimization, also known as black-box optimization or derivative-free methods, has been widely explored in many studies in recent years, especially in single-point gradient estimation. Although these methods have their own characteristics, rough experiments conducted by \citep{feedback_black_box} show that SPSA outperforms other methods.

Zeroth-Order Optimizer is widely used in situations where the objective function is difficult to express explicitly, or its gradient is impossible to obtain or extremely expensive to calculate. For example, derivative-free distributed algorithms have been proposed in ~\citep{distributed_ZO1, distributed_ZO2} to address nonconvex multi-agent optimization problems. A number of zero-order optimization methods, including ZO-BCD\citep{black_box_adversial_ZO2}, ZOO\citep{black_box_adversial_ZO1}, ZO-signSGD \citep{black_box_adversial_ZO3}, and ZO-HessAware \citep{black_box_adversial_ZO4}, are widely employed to generate black-box adversarial examples in deep learning.

MeZO~\citep{mezo} is the first to apply the classical ZO-SGD algorithm for fine-tuning large language models (LLMs), achieving competitive performance while significantly reducing memory usage and GPU time. This has led to a surge of follow-up works~\citep{jiang2023zoadamu, zhao2024helene, sparseMeZO, MeZOsparsity, tang2024effectivelyzo, chen2024zo_lowrank, wangadvancementzo, chen2025towardszo, tan2025harmonyzo, chen2025memoryzo, sun2025tezo}, which collectively establish zero-order optimization (ZOO) as a promising direction for LLM fine-tuning. Recently, HiZOO~\citep{hizoo} further extends this line by being the first to incorporate diagonal Hessian information into ZOO for LLMs, enabling better handling of heterogeneous curvatures across different parameter dimensions and achieving improved convergence and performance.

\clearpage

\section{Detailed Convergence Analysis}
\label{app_convergence_proof}

\subsection{Properties of Rademacher random vector}

\begin{lemma}
Letting $x\in\RR^d$ be a constant vector and $\{u_i\}_{i=1}^N$ be $d$-dimensional  Rademacher random vectors, then it holds that
\begin{equation}
\EE\left[ \norm{\sum_{i=1}^{N}u_i u_i^\top x}^2 \right] 
=
N(N+d-1)\norm{x}^2.
\end{equation}
\end{lemma}
\begin{proof}It holds that
\begin{align*}
\EE\left[ \norm{\sum_{i=1}^{N}u_i u_i^\top x}^2 \right]
=& 
\EE\left[\sum_{i=1}^{N}\sum_{j=1}^{N} x^\top u_i u_i^\top u_ju_j^\top x \right]\\
=&
\EE\left[\sum_{j=1}^{N} x^\top u_ju_j^\top x \cdot \norm{u_j}^2 \right] + \sum_{i\neq j}\dotprod{\EE\left[u_iu_i^\top x \right], \EE\left[u_ju_j^\top x\right]}\\
=&\EE\left[\sum_{j=1}^{N} x^\top u_ju_j^\top x \cdot \norm{u_j}^2 \right] + \sum_{i\neq j}\dotprod{\EE\left[u_iu_i^\top x \right], \EE\left[u_ju_j^\top x\right]}\\
=& N d \norm{x}^2  + (N^2-N) \norm{x}^2\\
=& N(N+d-1)\norm{x}^2
\end{align*}
\end{proof}

\begin{lemma}
Letting $u_i$'s be Rademacher random vector, then it holds that
\begin{equation}\label{eq:LL}
	\sum_{i=1}^{N}\EE\left[\dotprod{\nabla L(\theta, \cB), u_i - \frac{1}{N}\sum_{j=1}^{N}u_j}^2\right] 
	=
	(N-1) \norm{\nabla L(\theta, \cB)}^2. 
\end{equation}
\end{lemma}
\begin{proof}
It holds that
\begin{align}
	&\sum_{i=1}^{N}\dotprod{\nabla L(\theta, \cB), u_i - \frac{1}{N}\sum_{j=1}^{N}u_j}^2 \\
	=& \sum_{i=1}^{N}\dotprod{\nabla L(\theta, \cB), u_i}^2 
	+ N\dotprod{\nabla L(\theta, \cB), \frac{1}{N} \sum_{j=1}^{N}u_j}^2 \\
	&- 2 \sum_{i=1}^{N} \dotprod{\nabla L(\theta, \cB), u_i} \cdot\dotprod{\nabla L(\theta, \cB), \frac{1}{N} \sum_{j=1}^{N}u_j}  \\
	=&
	\sum_{i=1}^{N}\dotprod{\nabla L(\theta, \cB), u_i}^2 
	-  \frac{1}{N}\dotprod{\nabla L(\theta, \cB),  \sum_{j=1}^{N}u_j}^2
\end{align}
Furthermore,
\begin{align*}
	\EE\left[\dotprod{\nabla L(\theta, \cB),  \sum_{j=1}^{N}u_j}^2\right]
	=& \sum_{i=1}^{N}\sum_{j=1}^{N} \dotprod{\nabla L(\theta, \cB), u_i} \dotprod{\nabla L(\theta, \cB), u_j}\\
	=&
	\sum_{i=1}^{N} \EE\left[\dotprod{\nabla L(\theta, \cB), u_i}^2\right] + \sum_{i\neq j} \EE\left[\dotprod{\nabla L(\theta, \cB), u_i} \dotprod{\nabla L(\theta, \cB), u_j}\right]\\
	=&
	\sum_{i=1}^{N} \EE\left[\dotprod{\nabla L(\theta, \cB), u_i}^2\right] \\
	=& dN \norm{\nabla L(\theta, \cB)}^2. 
\end{align*}
Thus,
\begin{align*}
\sum_{j=1}^{N}\EE\left[\dotprod{\nabla L(\theta, \cB), u_i - \frac{1}{N}\sum_{j=1}^{N}u_j}^2 \right]
= 
\left(1 - \frac{1}{N}\right) \sum_{i=1}^{N} \EE\left[\dotprod{\nabla L(\theta, \cB), u_i}^2\right] 
=
(N-1) \norm{\nabla L(\theta, \cB)}^2.
\end{align*}
\end{proof}

\subsection{Properties of $g_t$ and $\sigma_t$}
By the Taylor's expansion, we have
\begin{equation}\label{eq:L_decom}
	L(\theta + \epsilon u_i, \cB ) = L(\theta) + \epsilon\dotprod{\nabla L(\theta, \cB), u_i} + \alpha(\theta, \epsilon u_i, \cB)
\end{equation}
where $\alpha(\theta, \epsilon u_i, \cB) \triangleq L(\theta + \epsilon u_i, \cB ) - \left( L(\theta) + \epsilon\dotprod{\nabla L(\theta, \cB), u_i} \right)$.
By the $\cL$-smoothness, we can obtain that
\begin{equation}\label{eq:alp}
	|\alpha(\theta, \epsilon u_i, \cB)| \leq \frac{\cL \epsilon^2}{2}\norm{u_i}^2 = \frac{d\cL\epsilon^2}{2},
\end{equation}
where the last equality is because $u_i\in \{-1,+1\}$.

\begin{lemma}\label{lem:sig}
Letting the estimated variance $\sigma_t$ defined in \eqref{eq:sigma},
then, it holds that
\begin{equation}\label{eq:sig1}
\EE\left[\sigma^2\right] = \epsilon^2\cdot \norm{\nabla L(\theta, \cB)}^2 + \zeta,
\end{equation}
with 
\begin{align*}
	\zeta &= \frac{1}{N-1}\sum_{i=1}^{N}  \Big(\alpha(\theta, \epsilon u_i, \cB) - \frac{1}{N}\sum_{j=1}^{N}\alpha(\theta, \epsilon u_j, \cB)\Big)^2 \\
	&+ 
	\epsilon\frac{1}{N-1}\sum_{i=1}^{N} \dotprod{\nabla L(\theta, \cB), u_i - \frac{1}{N}\sum_{j=1}^{N}u_j}\cdot \Big(\alpha(\theta, \epsilon u_i, \cB)- \frac{1}{N}\sum_{j=1}^{N}\alpha(\theta, \epsilon u_j, \cB)\Big),
\end{align*}
and
\begin{align*}
	|\zeta| \leq \frac{Nd^2\cL^2\epsilon^4}{2} + 2dN\cL\epsilon^3\cdot \norm{\nabla L(\theta, \cB)}.
\end{align*}
\end{lemma}
\begin{proof}
First, we have
\begin{align*}
	&L(\theta + \epsilon u_i, \cB ) - \frac{1}{N}\sum_{j=1}^{N} L(\theta + \epsilon u_j, \cB )
	\\
	\stackrel{\eqref{eq:L_decom}}{=}&
	L(\theta) + \epsilon\dotprod{\nabla L(\theta, \cB), u_i} + \alpha(\theta, \epsilon u_i, \cB) - \frac{1}{N}\sum_{j=1}^{N} \left( L(\theta) + \epsilon\dotprod{\nabla L(\theta, \cB), u_j} + \alpha(\theta, \epsilon u_j, \cB) \right)\\
	=& \epsilon \dotprod{\nabla L(\theta, \cB), u_i - \frac{1}{N}\sum_{j=1}^{N}u_j} + \alpha(\theta, \epsilon u_i, \cB) - \frac{1}{N}\sum_{j=1}^{N}\alpha(\theta, \epsilon u_j, \cB).
\end{align*}

\begin{align*}
	&\sum_{i=1}^{N} \left(L(\theta + \epsilon u_i, \cB ) - \frac{1}{N}\sum_{j=1}^{N} L(\theta + \epsilon u_j, \cB )\right)^2\\
	=&
	\sum_{i=1}^{N}\left( \epsilon \dotprod{\nabla L(\theta, \cB), u_i - \frac{1}{N}\sum_{j=1}^{N}u_j} + \beta_i \right)^2\\
	=&
	\sum_{i=1}^{N}\left( \epsilon^2 \dotprod{\nabla L(\theta, \cB), u_i - \frac{1}{N}\sum_{j=1}^{N}u_j}^2 + \beta_i^2 + \epsilon\dotprod{\nabla L(\theta, \cB), u_i - \frac{1}{N}\sum_{j=1}^{N}u_j} \beta_i \right),
\end{align*}
where we use $\beta_i \triangleq \alpha(\theta, \epsilon u_i, \cB) - \frac{1}{N}\sum_{j=1}^{N}\alpha(\theta, \epsilon u_j, \cB)$.

Furthermore, it holds that
\begin{align*}
	\sum_{i=1}^{N} \beta_i^2  
	= \left(1 -\frac{1}{N}\right)\sum_{i=1}^{N}  \alpha^2(\theta, \epsilon u_i, \cB)
	\leq \frac{Nd^2\cL^2\epsilon^4}{4},
\end{align*}
and
\begin{align*}
	\sum_{i=1}^{N}\left|\dotprod{\nabla L(\theta, \cB), u_i - \frac{1}{N}\sum_{j=1}^{N}u_j} \beta_i\right| 
	\leq 
	N\norm{\dotprod{\nabla L(\theta, \cB), u_i - \frac{1}{N}\sum_{j=1}^{N}u_j}  }\norm{\beta_i}
	\stackrel{\eqref{eq:alp}}{\leq}
	dN\cL\epsilon^2 \norm{\nabla L(\theta, \cB)}.
\end{align*}

Therefore, we can obtain that
\begin{align*}
	\EE\left[\sigma^2\right]
	=& \frac{1}{N-1} \sum_{i=1}^{N}\left( \epsilon^2 \dotprod{\nabla L(\theta, \cB), u_i - \frac{1}{N}\sum_{j=1}^{N}u_j}^2 + \beta_i^2 + \epsilon\dotprod{\nabla L(\theta, \cB), u_i - \frac{1}{N}\sum_{j=1}^{N}u_j} \beta_i \right)\\
	\stackrel{\eqref{eq:LL}}{=}&
	\epsilon^2\cdot \norm{\nabla L(\theta, \cB)}^2 +  \zeta
\end{align*}
Finally, we have
\begin{align*}
	|\zeta| 
	\leq& 
	\frac{1}{N-1} \sum_{i=1}^{N} \beta_i^2 + \epsilon \frac{1}{N-1} \sum_{i=1}^{N}\left|\dotprod{\nabla L(\theta, \cB), u_i - \frac{1}{N}\sum_{j=1}^{N}u_j} \beta_i\right| \\
	\leq&
	\frac{Nd^2\cL^2\epsilon^4}{2} + 2dN\cL\epsilon^3 \cdot \norm{\nabla L(\theta, \cB)}.
\end{align*}
\end{proof}

\begin{lemma}
Letting the stochastic gradient estimation $g_t$ defined in \eqref{eq:g_hat_t_def}, then it satisfies the following properties
\begin{equation}\label{eq:gt_decom}
	g_t= \frac{1}{N} \sum_{i=1}^{N}u_i u_i^\top \nabla L(\theta_t , \cB_t) + \tau_t
\end{equation}
with
\begin{equation}\label{eq:tau}
	\tau_t = \frac{1}{N}\sum_{i=1}^{N} \epsilon^{-1} \alpha(\theta, \epsilon u_i, \cB)\cdot u_i \mbox{ and } \EE\left[\tau_t^2\right] \leq \frac{d^3 \cL^2 \epsilon^2}{4N}.
\end{equation}
\end{lemma}
\begin{proof}
	By the definition of $g_t$ in \eqref{eq:g_hat_t_def}, we can obtain that
\begin{align*}
	g_t 
	\stackrel{\eqref{eq:g_hat_t_def}}{=}& \frac{1}{N}\sum_{i=1}^{N}\frac{L(\theta + \epsilon u_i, \cB) - L(\theta , \cB)}{\epsilon} u_i \\
	\stackrel{\eqref{eq:L_decom}}{=}& \frac{1}{N} \sum_{i=1}^{N}u_i u_i^\top \nabla L(\theta , \cB) + \frac{1}{N}\sum_{i=1}^{N} \epsilon^{-1} \alpha(\theta, \epsilon u_i, \cB)\cdot u_i\\
	=& \frac{1}{N} \sum_{i=1}^{N}u_i u_i^\top \nabla L(\theta , \cB) + \tau_t.
\end{align*}
Furthermore,
\begin{align*}
	\EE\left[\tau^2\right] 
	=& \frac{1}{N^2}\sum_{i=1}^{N}\EE\left[\left( \epsilon^{-1} \alpha(\theta, \epsilon u_i, \cB)\right)^2\cdot \norm{u_i}^2 \right]\\
	=& \frac{d\epsilon^{-2}}{N} \EE\left[\alpha^2(\theta, \epsilon u_j, \cB)\right]
	\stackrel{\eqref{eq:alp}}{\leq} \frac{d^3 \cL^2 \epsilon^2}{4N}.
\end{align*}
\end{proof}
\begin{lemma}\label{lem:gt_norm}
Letting the stochastic gradient estimation $g_t$ defined in \eqref{eq:g_hat_t_def}, then its norm satisfies that 
\begin{equation}\label{eq:gt_norm1}
	\EE\left[\norm{g_t}^2\right] 
	=
	\frac{N+d-1}{N} \norm{\nabla L(\theta_t , \cB_t)}^2
	+ 
	\gamma_t,
\end{equation}
with 
\begin{align*}
	\gamma_t =& \frac{d\epsilon^{-2}}{N} \EE\left[\alpha^2(\theta, \epsilon u_j, \cB)\right]
	+\frac{d \epsilon^{-1}}{N} \EE\left[ u_i^\top \nabla L(\theta , \cB)  \alpha(\theta, \epsilon u_i, \cB) \right]\\ 
	&+ \frac{(N-1)\epsilon^{-1}}{N} \dotprod{\nabla L(\theta , \cB), \EE\left[ \alpha(\theta, \epsilon u_j, \cB)\cdot u_j \right]}.
\end{align*}
Furthermore, it hols that
	\begin{align*}
		|\gamma_t| \leq \frac{d^3 \cL^2 \epsilon^2}{4N} + d^2 \cL \norm{\nabla L(\theta_t,\cB_t)} \cdot \epsilon.
	\end{align*}
\end{lemma}
\begin{proof}
First, we have
\begin{align*}
\norm{g_t}^2 
=&
\frac{1}{N^2}\norm{\sum_{i=1}^{N}u_i u_i^\top \nabla L(\theta , \cB)}^2 + \frac{1}{N^2}\sum_{i=1}^{N}\left( \epsilon^{-1} \alpha(\theta, \epsilon u_i, \cB)\right)^2\cdot \norm{u_i}^2 \\
&+ \frac{1}{N^2}\sum_{i=1}^{N}\sum_{j=1}^{N} \dotprod{u_i u_i^\top \nabla L(\theta , \cB), \epsilon^{-1} \alpha(\theta, \epsilon u_i, \cB)\cdot u_i}\\
=&
\frac{1}{N^2}\norm{\sum_{i=1}^{N}u_i u_i^\top \nabla L(\theta , \cB)}^2 + \frac{d}{N^2}\sum_{i=1}^{N}\left( \epsilon^{-1} \alpha(\theta, \epsilon u_i, \cB)\right)^2
\\
&+ \frac{1}{N^2}\sum_{i=1}^{N}\sum_{j=1}^{N} \dotprod{u_i u_i^\top \nabla L(\theta , \cB), \epsilon^{-1} \alpha(\theta, \epsilon u_j, \cB)\cdot u_j}.
\end{align*}
We also have
\begin{align*}
\EE\left[ \norm{\sum_{i=1}^{N}u_i u_i^\top \nabla L(\theta , \cB)}^2 \right] 
= N(N+d-1) \norm{ \nabla L(\theta , \cB) }^2.
\end{align*}
Furthermore, it holds that
\begin{align*}
&\EE\left[ \sum_{i=1}^{N}\sum_{j=1}^{N} \dotprod{u_i u_i^\top \nabla L(\theta , \cB), \epsilon^{-1} \alpha(\theta, \epsilon u_j, \cB)\cdot u_j} \right]\\
=& 
\EE\left[\sum_{i=1}^{N} \dotprod{u_i u_i^\top \nabla L(\theta , \cB),  \alpha(\theta, \epsilon u_i, \cB)\cdot u_i}\right]
+\EE\left[ \sum_{i\neq j}^{N} \dotprod{u_i u_i^\top \nabla L(\theta , \cB), \epsilon^{-1} \alpha(\theta, \epsilon u_j, \cB)\cdot u_j} \right]\\
=&
Nd \epsilon^{-1} \EE\left[ u_i^\top \nabla L(\theta , \cB)  \alpha(\theta, \epsilon u_i, \cB) \right] 
+ N(N-1)\epsilon^{-1} \dotprod{\nabla L(\theta , \cB), \EE\left[ \alpha(\theta, \epsilon u_j, \cB)\cdot u_j \right]}.
\end{align*}

Thus, 
\begin{align*}
\EE\left[\norm{g_t}^2\right] 
=& 
\frac{N+d-1}{N} \norm{\nabla L(\theta , \cB)}^2
+ 
\frac{d\epsilon^{-2}}{N} \EE\left[\alpha^2(\theta, \epsilon u_j, \cB)\right]\\
&+\frac{d \epsilon^{-1}}{N} \EE\left[ u_i^\top \nabla L(\theta , \cB)  \alpha(\theta, \epsilon u_i, \cB) \right] 
+ \frac{(N-1)\epsilon^{-1}}{N} \dotprod{\nabla L(\theta , \cB), \EE\left[ \alpha(\theta, \epsilon u_j, \cB)\cdot u_j \right]}.
\end{align*}

Finally,
\begin{align*}
	|\gamma_t| 
	\leq&  
	\frac{d\epsilon^{-2}}{N} \left|\EE\left[\alpha^2(\theta, \epsilon u_j, \cB)\right]\right|
	+\frac{d \epsilon^{-1}}{N} \left|\EE\left[ u_i^\top \nabla L(\theta , \cB)  \alpha(\theta, \epsilon u_i, \cB) \right]\right| \\
	&
	+ \frac{(N-1)\epsilon^{-1}}{N} \left|\dotprod{\nabla L(\theta , \cB), \EE\left[ \alpha(\theta, \epsilon u_j, \cB)\cdot u_j \right]}\right|\\
	\stackrel{\eqref{eq:alp}}{\leq}& \frac{d\epsilon^{-2}}{N} \cdot \frac{d^2\cL^2\epsilon^4}{4} + \frac{d\epsilon^{-1}}{N} \norm{\nabla L(\theta_t, \cB_t)} \cdot \frac{d\cL\epsilon^2}{2}
	+ \epsilon^{-1} \norm{\nabla L(\theta_t,\cB_t)} \cdot \frac{d\cL\epsilon^2}{2}\cdot d^{1/2}\\
	=&\frac{d^3 \cL^2 \epsilon^2}{4N} + \left(\frac{d^2\cL \epsilon}{2N}+ \frac{d^{3/2} \cL \epsilon}{2}\right) \norm{\nabla L(\theta_t,\cB_t)}\\
	\leq&
	\frac{d^3 \cL^2 \epsilon^2}{4N} + d^2 \cL \norm{\nabla L(\theta_t,\cB_t)} \cdot \epsilon,
\end{align*}
which concludes the proof.
\end{proof}

\subsection{Proof of Proposition~\ref{prop:nsgd}}

 \begin{proof}
    Lemma~\ref{lem:sig} and Lemma~\ref{lem:sig} imply the result of Proposition~\ref{prop:nsgd}.
 \end{proof}

\subsection{Proof of Theorem~\ref{thm:fzoo_convergence}}
\begin{proof}
First, we have
	\begin{equation}\label{eq:dec}
	\EE\left[\dotprod{\nabla L(\theta_t), -\tau_t}\right]
	\leq \frac{\norm{\nabla L(\theta_t)}^2 + \EE\norm{\tau_t}^2}{2}
	\stackrel{\eqref{eq:tau}}{\leq}
	\frac{\norm{\nabla L(\theta_t)}^2}{2}
	+\frac{d^3 \cL^2 \epsilon^2}{8N},
	\end{equation}
	where the first inequality is because of the Cauchy's inequality.
	
	It also holds that
	\begin{equation}\label{eq:eps}
	d\cL^2\sigma_t^{-2}\eta^2 \epsilon \cdot \norm{\nabla L(\theta_t, \cB_t)}
	\leq \frac{2d\cL\sigma_t^{-2}\eta^2 \norm{\nabla L(\theta_t, \cB_t)}^2}{N} + \frac{Nd\cL^3\sigma_t^{-2}\eta^2\epsilon^2}{2}.
	\end{equation}
Thus, by the $\cL$-smoothness of $\nabla L(\theta)$, we have
\begin{align*}
	\EE\left[L(\theta_{t+1})\right]
	\leq& 
		L(\theta_t) - \eta \sigma_t^{-1} \EE\left[\dotprod{\nabla L(\theta_t), g_t}\right] + \frac{\cL\sigma_t^{-2}\eta^2}{2} \EE\left[\norm{g_t}^2\right]\\
	\stackrel{\eqref{eq:gt_decom}}{=}&
	L(\theta_t) - \eta \sigma_t^{-1} \EE\left[\dotprod{\nabla L(\theta_t), \frac{1}{N}\sum_{i=1}^{N}u_iu_i^\top  \nabla L(\theta_t, \cB_t) + \tau_t}\right] + \frac{\cL\sigma_t^{-2}\eta^2}{2} \EE\left[\norm{g_t}^2\right]\\
	=& 
	L(\theta_t) - \eta \sigma_t^{-1} \norm{\nabla L(\theta_t)}^2 +\eta \sigma_t^{-1} \EE\left[\dotprod{\nabla L(\theta_t), -\tau_t}\right]
	+ \frac{\cL\sigma_t^{-2}\eta^2}{2} \EE\left[\norm{g_t}^2\right]\\
	\stackrel{\eqref{eq:dec}\eqref{eq:gt_norm}}{\leq}&
	L(\theta_t) - \frac{\eta \sigma_t^{-1}}{2} \norm{\nabla L(\theta_t)}^2+\frac{\eta \sigma_t^{-1} d^3 \cL^2 \epsilon^2}{8N}
	+ \frac{N+d-1}{N} \cdot\frac{\cL\sigma_t^{-2} \eta^2}{2} \norm{\nabla L(\theta_t, \cB_t)}^2 \\
	&+ \frac{d^3\cL^3\sigma_t^{-2}\eta^2 \epsilon^2}{8N} + d\cL^2\sigma_t^{-2}\eta^2 \epsilon \cdot \norm{\nabla L(\theta_t, \cB_t)}\\
	\stackrel{\eqref{eq:eps}}{\leq}& 
	L(\theta_t) - \frac{\eta \sigma_t^{-1}}{2} \norm{\nabla L(\theta_t)}^2+\frac{\eta \sigma_t^{-1} d^3 \cL^2 \epsilon^2}{8N}
	+ \frac{2d \cL\sigma_t^{-2} \eta^2}{N} \cdot \norm{\nabla L(\theta_t, \cB_t)}^2 \\
	&+ \frac{d^3\cL^3\sigma_t^{-2}\eta^2 \epsilon^2}{8N} + \frac{2d\cL\sigma_t^{-2}\eta^2 \norm{\nabla L(\theta_t, \cB_t)}^2}{N} + \frac{Nd\cL^3\sigma_t^{-2}\eta^2\epsilon^2}{2}
	\\
	\stackrel{\eqref{eq:sigma}}{\leq}&
	L(\theta_t) - \frac{\eta \sigma_t^{-1}}{2} \norm{\nabla L(\theta_t)}^2 + \frac{4d\cL \sigma_t^{-2}\eta^2}{N} \norm{\nabla L(\theta_t)}^2 + \frac{4d\cL \sigma_t^{-2}\eta^2 \ccV^2}{N} \\
	& +\frac{d^3\cL^3\sigma_t^{-2}\eta^2 \epsilon^2}{8N} + \frac{Nd\cL^3\sigma_t^{-2}\eta^2\epsilon^2}{2}\\
	\leq& L(\theta_t) - \frac{\eta \sigma_t^{-1}}{4} \norm{\nabla L(\theta_t)}^2 +  \frac{4d\cL \sigma_t^{-2}\eta^2 \ccV^2}{N} 
	+\frac{d^3\cL^3\sigma_t^{-2}\eta^2 \epsilon^2}{8N} + \frac{Nd\cL^3\sigma_t^{-2}\eta^2\epsilon^2}{2},
\end{align*}
where the last inequality is because of step size satisfies $\eta \sigma_t^{-1} \leq \frac{N}{16d\cL}$.

We can represent above equation as follows:
\begin{align*}
	\frac{\eta\sigma_t^{-1}}{4} \EE\left[ \norm{\nabla L(\theta_t)}^2 \right] 
	\leq \EE\left[ L(\theta_t) - L(\theta_{t+1}) \right] +   \frac{4d\cL \sigma_t^{-2}\eta^2 \ccV^2}{N} 
	+\frac{d^3\cL^3\sigma_t^{-2}\eta^2 \epsilon^2}{8N} + \frac{Nd\cL^3\sigma_t^{-2}\eta^2\epsilon^2}{2}.
\end{align*}
Telescoping above equation, we can obtain that
\begin{align*}
	\frac{1}{T}\sum_{t=1}^{T}\frac{\eta\sigma_t^{-1}}{4} \EE\left[ \norm{\nabla L(\theta_t)}^2 \right] 
	\leq 
	\frac{L(\theta_1) - L(\theta^*)}{T} + \sum_{t=1}^{T}\left(\frac{4d\cL \sigma_t^{-2}\eta^2 \ccV^2}{TN} 
	+\frac{d^3\cL^3\sigma_t^{-2}\eta^2 \epsilon^2}{8TN} + \frac{Nd\cL^3\sigma_t^{-2}\eta^2\epsilon^2}{2T}\right)
\end{align*}
\begin{align*}
\frac{1}{T}\sum_{t=1}^{T} \EE\left[ \norm{\nabla L(\theta_t)}^2 \right]
\leq& 
\frac{1}{T}\sum_{t=1}^{T}\frac{\sigma_t^{-1}}{ \sigma_*^{-1}} \EE\left[ \norm{\nabla L(\theta_t)}^2 \right] \\
\leq&
\frac{4\sigma_*(L(\theta_1) - L(\theta^*))}{\eta T } + 4\sigma_* \sum_{t=1}^{T}\left(\frac{4d \cL \sigma_t^{-2}\eta \ccV^2}{TN} 
+\frac{d^3\cL^3\sigma_t^{-2}\eta \epsilon^2}{8TN} + \frac{Nd\cL^3\sigma_t^{-2}\eta\epsilon^2}{2T}\right)
\end{align*}

By setting 
\begin{align*}
	\eta = \left(\frac{L(\theta_1) - L(\theta^*)}{4d\cL \ccV^2 \sum_{t=1}^{T} \sigma_t^{-2}}\right)^{1/2},
\end{align*}
we can obtain that
\begin{align*}
\frac{1}{T}\sum_{t=1}^{T} \EE\left[ \norm{\nabla L(\theta_t)}^2 \right]
\leq& 
\frac{64\sigma_* \left(d\cL(\theta_1) - L(\theta^*) \sum_{t=1}^{T} \sigma_t^{-2} \right)^{1/2}}{T}\\
&+ \frac{4\sigma_* \epsilon^2}{T} \left(\frac{d^3\cL^3}{8N} + \frac{Nd\cL^3}{2}\right)  \left(\frac{\Big(L(\theta_1) - L(\theta^*)\Big) \sum_{t=1}^{T} \sigma_t^{-2}}{4d\cL \ccV^2 }\right)^{1/2}
\end{align*}

\begin{align*}
	\eta \sigma_t^{-1} \leq \frac{N}{16d\cL}
\end{align*}
Accordingly, we can obtain that
\begin{align*}
\EE\left[L(\theta_{t+1})\right] 
\leq L(\theta_t) - \frac{\eta \sigma_t^{-1}}{4} \norm{\nabla L(\theta_t)}^2 +  \frac{4d\cL \sigma_t^{-2}\eta^2 \ccV^2}{N} 
+\frac{d^3\cL^3\sigma_t^{-2}\eta^2 \epsilon^2}{8N} + \frac{Nd\cL^3\sigma_t^{-2}\eta^2\epsilon^2}{2}.
\end{align*}

\end{proof}

\clearpage

\section{FZOO Variants}
\label{app_MeZO_Hessian_variants}

\subsection{FZOO-R}
\label{HiZOO_L}

\begin{figure}[h]

\centering
\scalebox{1.0}{
\includegraphics[width=1.0\linewidth]{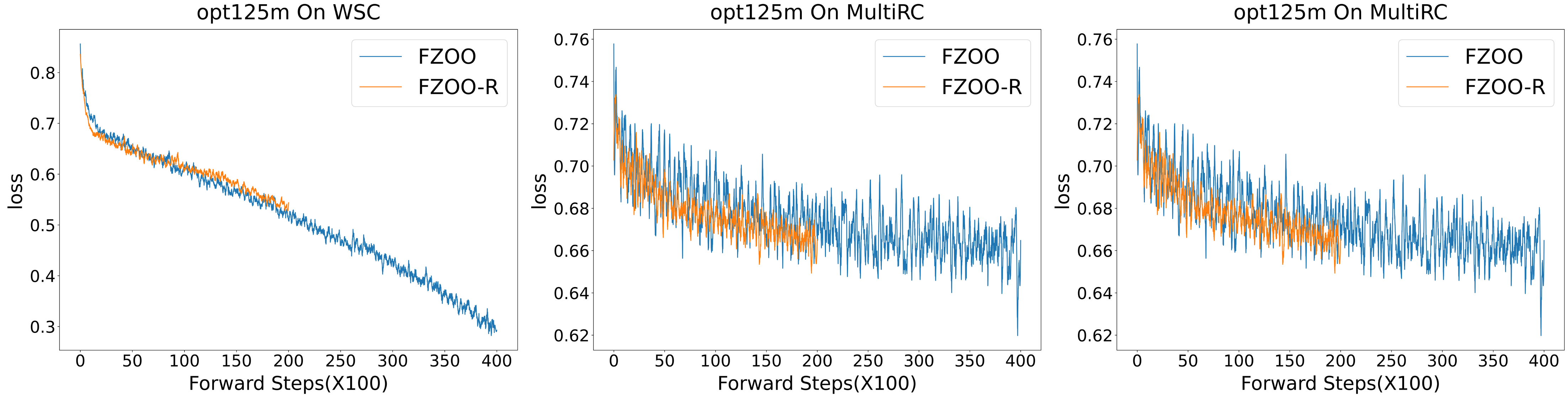}}
\vspace{-3mm}
\caption{Loss curves on opt-125m between FZOO and FZOO-R.}
\label{fig:ada_llama_loss}

\end{figure}

Although FZOO accelerates training convergence through adaptive step size adjustment, each update requires multiple forward passes. To reduce the computational overhead per update, we propose a loss reuse variant FZOO-R. Instead of performing the full recommended number of forward passes for each update, we halve it. FZOO-R estimates the gradient direction solely based on these few forward passes, while reusing loss values from the previous update to adjust the step size. The detailed procedure is described in Algorithm ~\ref{alg:FZOO-R}. We visualize the loss curves of FZOO-R and FZOO in Figure ~\ref{fig:ada_llama_loss}. On most datasets, FZOO-R achieves comparable convergence to FZOO while requiring only half the update time per step.

\begin{algorithm}[t]
\begin{algorithmic}[1]
\Require parameters $\theta \in \mathbb{R}^d$, loss $L : \mathbb{R}^d \rightarrow \mathbb{R}$, step budget $T$, perturbation scale $\epsilon$, batch size $N$, learning rate schedule $\{\eta_t\}$

\State Initialize $l_{-1} \gets \emptyset$  \Comment{To store previous losses}

\For{$t=1,...,T$}
    \State $\ell$, $\theta$, $seeds$ $\leftarrow$ BatchPerturbParameters($\theta$, $\epsilon$, $N$)
    \If{$l_{-1}$ is not empty}
       \State $std$ $\leftarrow$ standard deviation of \text{concat}($\ell, l_{-1}$)
    \Else
        \State $std$ $\leftarrow$ standard deviation of $\ell$
    \EndIf
    
    \State $projected\_grad \leftarrow (\ell - \mathcal{L}(\theta; \mathcal{B})) / (N * std)$
    \State $l_{-1} \gets \ell $ \Comment{Save the current loss}
    \State BatchUpdateParameter($projected\_grad$, $seeds$, $\theta$, ${\eta_t}$)
\EndFor

\Function{BatchPerturbParameter}{$\theta$, $\epsilon$, $N$}
    \State Sample batch $ \mathcal{B} \subset \mathcal{D}$; obtain input $X$ and first layer weights $W^{(1)}$
    \State Initialize random seeds $seeds \leftarrow \{s_1, \dots, s_N\}$
    \State Generate perturbation vectors $u \in \{\pm 1\}^{N \times d}$ using $seeds$
    \State Compute unperturbed activations $F^{(1)} \leftarrow W^{(1)}X$
    \State Compute perturbed activations: $Y^{(1)}_i \leftarrow F^{(1)} + \epsilon(u_i \odot X),\, i=1,\dots,N$
    \State Concatenate activations $Y^{(1)} \leftarrow [Y^{(1)}_1;\dots;Y^{(1)}_N]$
    \For{$j=2,3,\dots$}
        \State $F^{(j)} \leftarrow W^{(j)}Y^{(j-1)}$ \Comment{Compute unperturbed activations at layer $j$}
        \State $P^{(j)} \leftarrow \epsilon(u \odot Y^{(j-1)})$ \Comment{Compute perturbations at layer $j$}
        \State $Y^{(j)} \leftarrow F^{(j)} + P^{(j)}$ \Comment{Compute perturbed activations in parallel at layer $j$}
    \EndFor
    \State $\ell \leftarrow \mathcal{L}(Y^{(final)}; \mathcal{B})$ \Comment{Compute final losses in parallel}
    \State \Return $\ell$, $\theta$, $seeds$
\EndFunction

\Function{BatchUpdateParameter}{$projected\_grad$, $seeds$, $\theta$, ${\eta}$}
    \For{$idx, s \in \texttt{enumerate}(seeds)$}
        \State Reset random number generator with seed $s$ 
        \Comment{For sampling $u$}
        \For{$\theta_i \in \theta$}
            \State $u \sim \text{Uniform}(\{+1, -1\})$
            \State $\theta_i \leftarrow \theta_i - \eta \ast  projected\_grad_{idx} \ast u$
        \EndFor
    \EndFor
\EndFunction
  \caption{FZOO-R}
  \label{alg:FZOO-R}
  \end{algorithmic}
\end{algorithm}

\subsection{FZOO without parallel}
\label{FZOO_without_parallel}

To facilitate controlled evaluation, we release a \emph{non‑parallel} variant of FZOO, allowing for a clearer comparison of the total step‑count reduction ratio against baselines such as MeZO, as illustrated in Algorithm~\ref{alg:FZOO-wop}.  
In this version, the perturbation noise is sampled from a Rademacher random vector ($\pm1$), ensuring that, apart from the removal of parallel optimization, all other components remain identical to the original FZOO.  
\textbf{Note.} Gaussian sampling benefits from dedicated optimized kernels, whereas PyTorch currently lacks a specialized routine for Rademacher sampling. Hence, generating Rademacher noise via  
\texttt{torch.randint(0,\ 2,\ size=param.data.size(),\ device=param.data.device,\ dtype=param.data.dtype)\ *\ 2\ -\ 1}  
is slower than Gaussian sampling.

\begin{algorithm}[t]
\begin{algorithmic}[1]
\Require parameters $\theta \in \mathbb{R}^d$, loss $L : \mathbb{R}^d \rightarrow \mathbb{R}$, step budget $T$, perturbation scale $\epsilon$, batch size $N$, learning rate schedule $\{\eta_t\}$

\For{$t=1,...,T$}
    \State $\ell$, $\theta$, $seeds$ $\leftarrow$ BatchPerturbParameters($\theta$, $\epsilon$, $N$)
    \State $std$ $\leftarrow$ standard deviation of $\ell$
    \State $projected\_grad \leftarrow (\ell - \mathcal{L}(\theta; \mathcal{B})) / (N * std)$
    \State BatchUpdateParameter($projected\_grad$, $seeds$, $\theta$, ${\eta_t}$)
\EndFor

\Function{BatchPerturbParameter}{$\theta$, $\epsilon$, $N$}
    \State Sample batch $ \mathcal{B} \subset \mathcal{D}$ 
    \For{$i=1,...,N$}
        \State Sample random seed $s$ and append to $seeds$
        \State $\theta$ $\leftarrow$ PerturbParameters($\theta$, $\epsilon$, $s$)
        \State Append $\mathcal{L}(\theta; \mathcal{B})$ to $\ell$ 
        \State $\theta$ $\leftarrow$ PerturbParameters($\theta$, -$\epsilon$, $s$)
        \Comment{Reset parameters before next forward pass}
    \EndFor
    \State \Return $\ell$, $\theta$, $seeds$
\EndFunction
\Function{PerturbParameter}{$\theta$, $\epsilon$, $s$}
    \State Reset random number generator with seed $s$  
    \Comment{For sampling $u$}
    \For{$\theta_i \in \theta$}
        \State $u \sim \text{Uniform}(\{+1, -1\})$
        \State $\theta_i \leftarrow \theta_i + \epsilon u$
        \Comment{Modify parameters in place}
    \EndFor
    \State \Return $\theta$
\EndFunction
\Function{BatchUpdateParameter}{$projected\_grad$, $seeds$, $\theta$, ${\eta}$}
    \For{$idx, s \in \texttt{enumerate}(seeds)$}
        \State Reset random number generator with seed $s$ 
        \Comment{For sampling $u$}
        \For{$\theta_i \in \theta$}
            \State $u \sim \text{Uniform}(\{+1, -1\})$
            \State $\theta_i \leftarrow \theta_i - \eta \ast  projected\_grad_{idx} \ast u$
        \EndFor
    \EndFor
\EndFunction
  \caption{FZOO without parallel}
  \label{alg:FZOO-wop}
  \end{algorithmic}
\end{algorithm}

\section{Experiments on LLMs}

\subsection{Detailed Experiments on RoBERTa-large}
\label{appendix_robert}

We perform \nameo\ experiments on the RoBERTa-large model using the hyperparameters listed in Table \ref{tab:roberta_hyper} . For learning rate scheduling and early stopping strategies, a constant learning rate is used in all \nameo\ experiments.

\begin{table*}[h]
\centering
\caption{The hyperparameter grids used for RoBERTa-large experiments. \nameo\ uses a constant learning rate schedule.}
\small
\begin{tabular}{lrc}
\toprule
Experiment & Hyperparameters & Values \\
\midrule
\name & Batch size & $64$ \\
& Learning rate & $1\mathrm{e}{-4} $ \\
& $\mu$ & $1\mathrm{e}{-4}$ \\
& Weight Decay & $0$ \\
\midrule
\name (prefix) & Batch size & $64$ \\
& Learning rate & $\{1\mathrm{e}{-2}, 1\mathrm{e}{-1} \}$ \\
& $\mu$ & $\{ 1\mathrm{e}{-2}, , 1\mathrm{e}{-1} \}$ \\
& Weight Decay & $0$ \\
& \# prefix tokens &$5$\\

\bottomrule
\end{tabular}

\label{tab:roberta_hyper}
\end{table*}
\begin{table*}[htbp]
  \centering
  \caption{Experiments on RoBERTa-large (350M parameters, k=512). For MeZO we report the results we reproduced.}
    \vspace{2pt}
    \scalebox{0.8}{
    \begin{tabular}{lccccccc}
    \toprule
    Task Type & \multicolumn{1}{c}{\textbf{SST-2}} & \multicolumn{1}{c}{\textbf{SST-5}} & \multicolumn{1}{c}{\textbf{SNLI}}  & \multicolumn{1}{c}{\textbf{MNLI}} & \multicolumn{1}{c}{\textbf{RTE}} & \multicolumn{1}{c}{\textbf{TREC}} & \multicolumn{1}{c}{\textbf{Average}}\\
    & \multicolumn{2}{c}{------ sentiment ------} & \multicolumn{3}{c}{------ natural language inference ------} & \multicolumn{1}{c}{--- topic ---}\\
    \midrule
    Zero-shot & 79.0  & 35.5  & 50.2  & 48.8  & 51.4  & 32.0 &49.5\\
        LP    &  91.3 & 51.7 & 80.9 & 71.5 & 73.1 & 89.4 &76.3\\
    \midrule
    FT  & 91.9 & 47.5 & 77.5 & 70.0 & 66.4 & 85.0 &73.1\\
    FT (prefix) &  91.9 & 47.7 & 77.2 & 66.5 & 66.6 & 85.7 &72.6\\
    \midrule
    MeZO  & 93.3 & 53.2 & 83.0 & 78.3 & 78.6 & 94.3 &80.1\\
    MeZO (prefix) & 93.3 & 53.6 & 82.9 & 75.6 & 77.2 & 88.2 &78.4\\
    \midrule
    HiZOO  & \textbf{95.5} & 53.2 & 82.6 & 77.7 & \textbf{80.0} & 94.6 &80.6\\
    HiZOO (prefix) & \textbf{96.1} & \textbf{54.2} & \textbf{85.7} & 79.7 & \textbf{77.3} & 93.9 &\textbf{81.2}\\
    \midrule
    FZOO  & 94.2  & \textbf{55.1}  & \textbf{84.6}  & \textbf{79.9}  & 79.1  & \textbf{95.6}  & \textbf{81.4}\\
    FZOO (prefix) & \textbf{96.1}  & 51.5  & 83.1  & \textbf{81}  & 76.5 & \textbf{95.8}  & 80.7\\
    \bottomrule
    \end{tabular}}
    \label{tab:roberta_k512}%
\end{table*}%

\begin{figure}[h]
\centering
\scalebox{1.0}{
\includegraphics[width=1.0\linewidth]{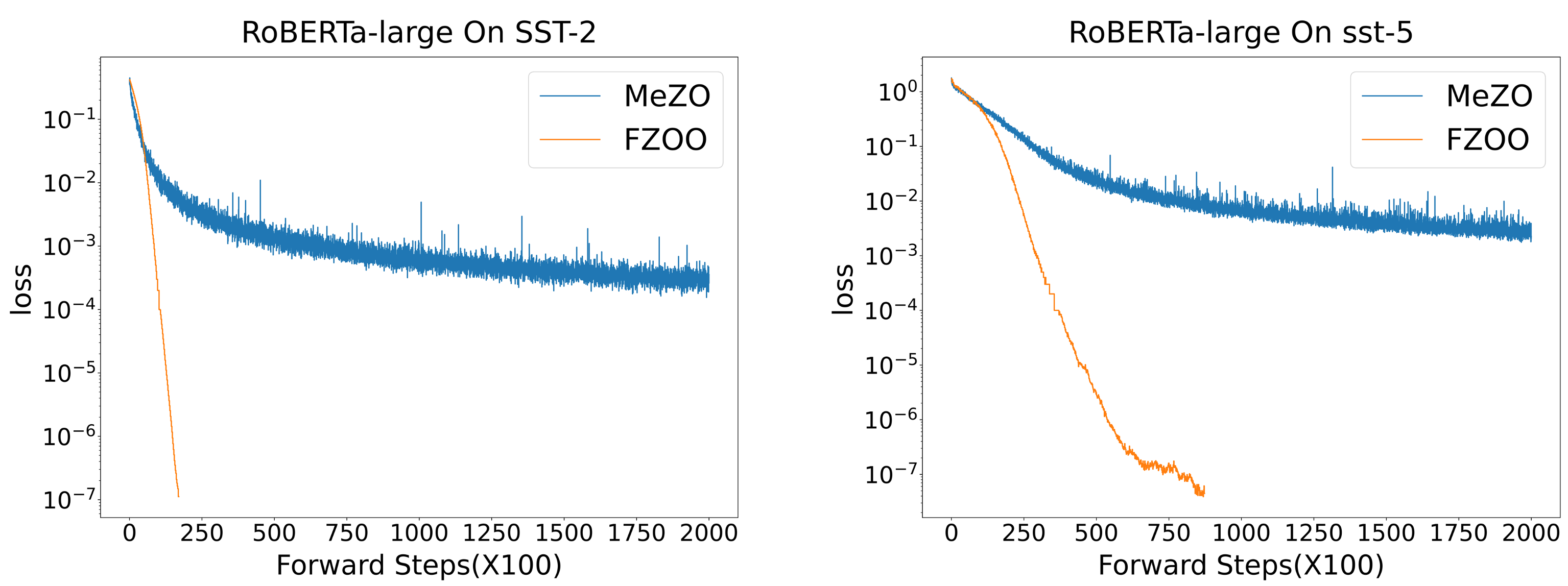}}
\caption{Loss curves on RoBERTa-large between MeZO and FZOO.}
\label{fig:app_roberta1}
\end{figure}

In Table \ref{tab:roberta_k512} we show the full experiment results. Additionally, we plot more loss curves to compare with MeZO. As shown in Figure \ref{fig:app_roberta1}, we can see that \name can greatly accelerate the training process over MeZO, which verifies the robustness of \name.

\subsection{More results on various LLMs}
\label{app:Llama3_phi2_opt}

\begin{figure}[t]
\centering
    \begin{minipage}{1.0\linewidth}
    \centering
    \scalebox{1.1}{
    \begin{minipage}{0.9\linewidth}
        \includegraphics[width=\linewidth]{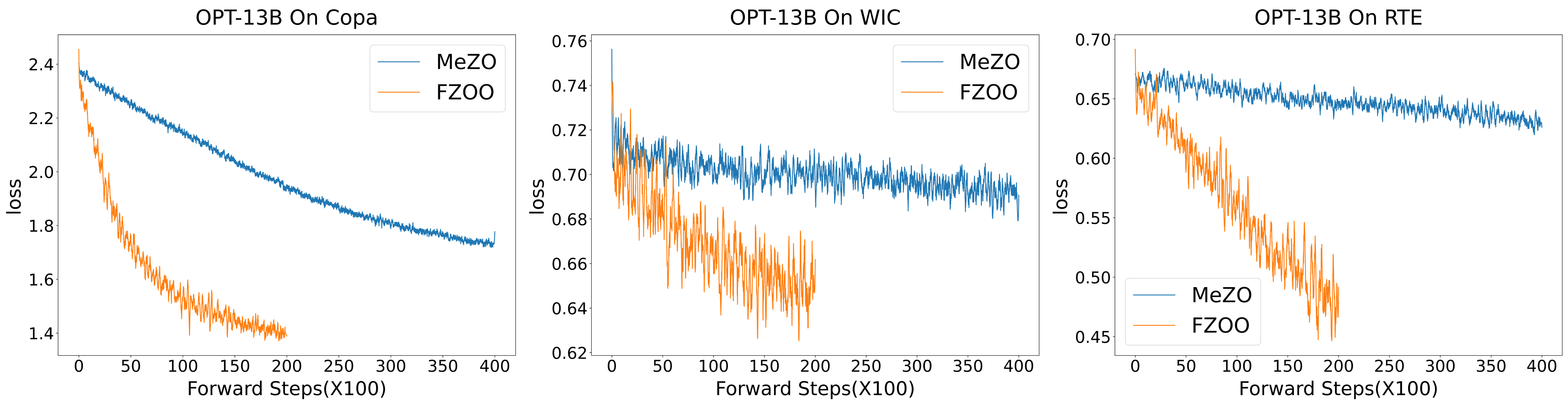}
    \end{minipage}
    }


    \scalebox{1.1}{
    \begin{minipage}{0.9\linewidth}
        \includegraphics[width=\linewidth]{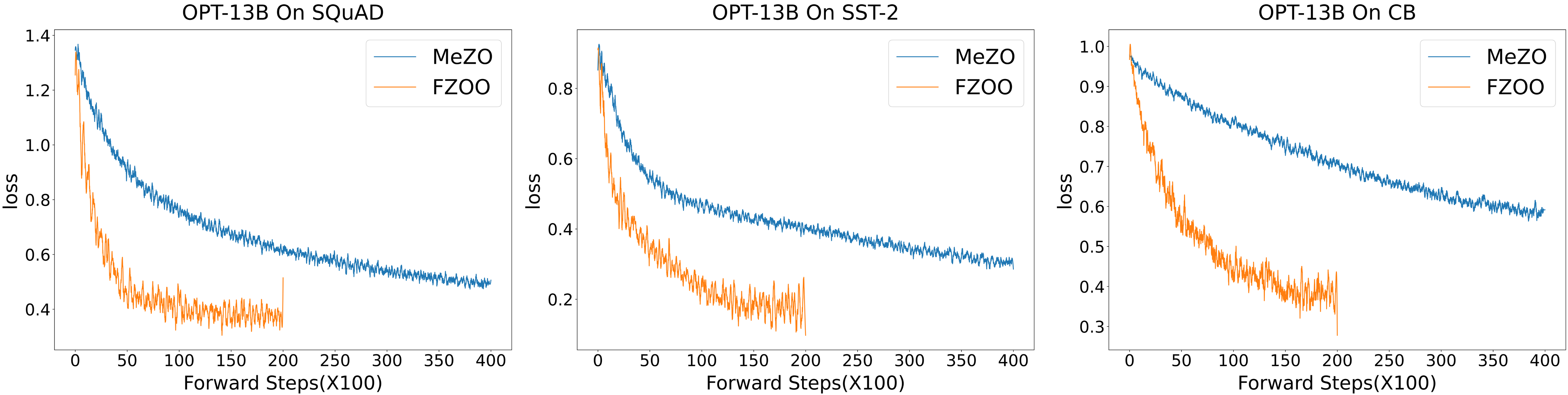}
    \end{minipage}
    }

    \vspace{-2mm}
    \caption{Loss curves on OPT.}
    \vspace{1mm}
    \label{fig:app_opt}
    \end{minipage}

\end{figure}

\begin{figure}[h]
\centering
    \begin{minipage}{1.0\linewidth}
    \centering
    \scalebox{1.1}{
    \includegraphics[width=0.9\linewidth]{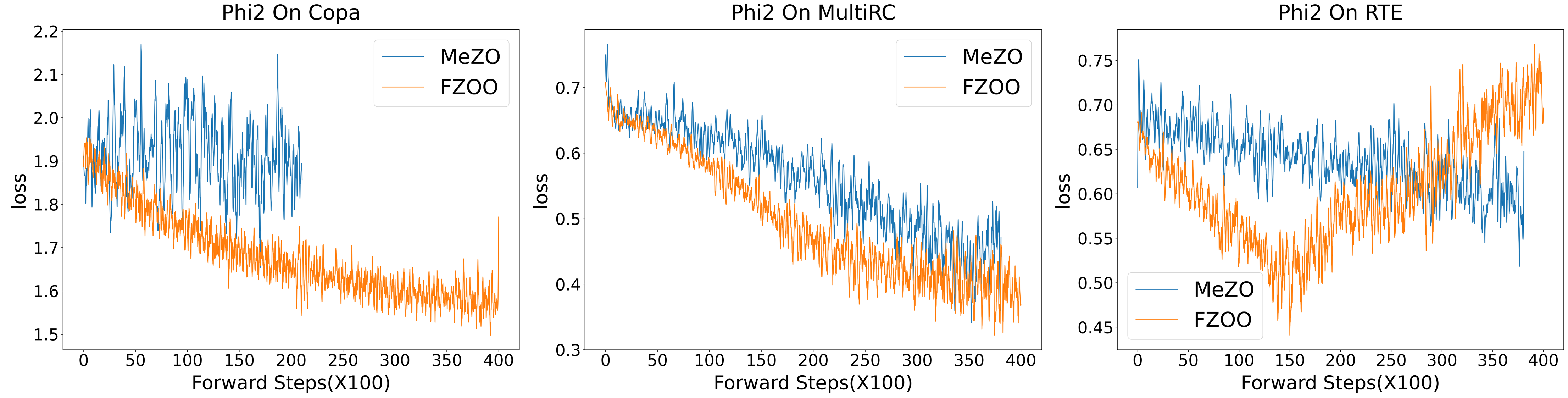}}
    \vspace{-7mm}
    \caption{Loss curves on Phi-2 between MeZO and FZOO.}
    \vspace{2mm}
    \label{fig:app_Phi2}
    \end{minipage}
\qquad
    \begin{minipage}{1.0\linewidth}
    \centering
    \scalebox{1.1}{
    \includegraphics[width=0.9\linewidth]{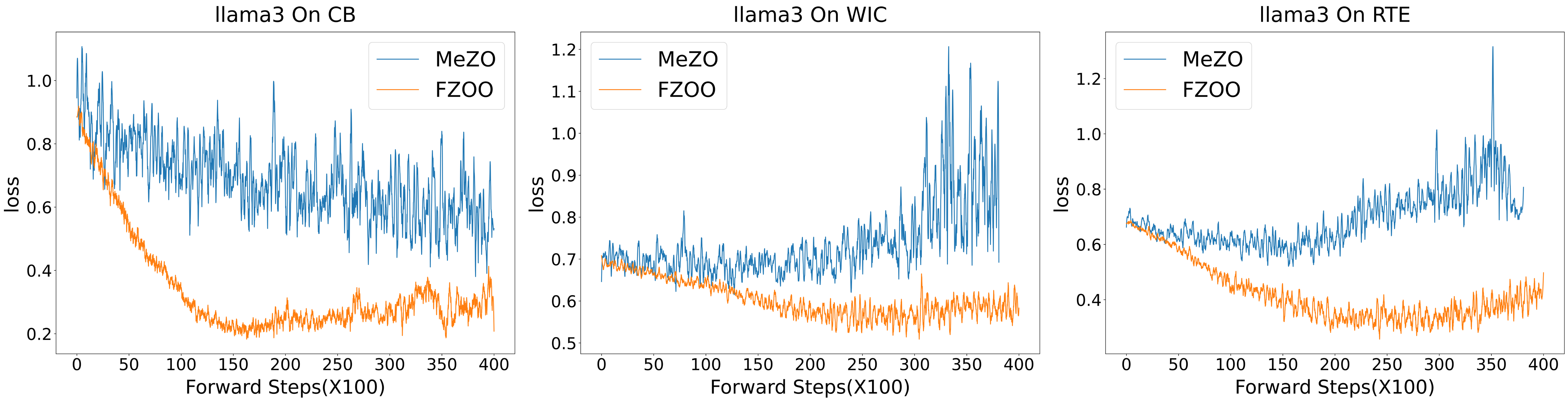}}
    \vspace{-6mm}
    \caption{Loss curves on Llama3 between MeZO and FZOO.}
    \label{fig:app_Llama3}
    \end{minipage}
\end{figure}

We use the hyperparameters in Table~\ref{tab:opt_hyper} for FZOO experiments on OPT. 
Full results for OPT-2.7B and OPT-125M are in  Table~\ref{tab:app_opt125m_2.7b}. \textbf{We also provide the relative loss curves of fine-tuning OPT-13B in Figure~\ref{fig:app_opt}.} We also provide several loss curves of fine-tuning Phi-2(2.7B) and Llama3(8B) in Figure~\ref{fig:app_Phi2} and Figure~\ref{fig:app_Llama3}.

\begin{table*}[h]
\caption{The hyperparameter grids used for OPT experiments. All weight decay is set to $0$. \name uses constant learning rates.}
    \centering
    \small
    \begin{tabular}{lrc}
    \toprule
    Experiment & Hyperparameters & Values \\
    \midrule
    \name & Batch size & $16$ \\
   & Learning rate & $\{1\mathrm{e}{-5}, 5\mathrm{e}{-5}, 1\mathrm{e}{-4}, 5\mathrm{e}{-4} \}$\\
    & $\mu$ & $1\mathrm{e}{-3}, 5\mathrm{e}{-4}, 1\mathrm{e}{-4}$ \\

    \midrule
    FT with Adam & Batch size & $8$ \\
    & Learning Rates & $\{1\mathrm{e}{-5}, 5\mathrm{e}{-5}, 8\mathrm{e}{-5} \}$\\
    \bottomrule
    \end{tabular}
    
    \label{tab:opt_hyper}
    \end{table*}

\begin{table*}[h]
\centering
\caption{Experiments on OPT-2.7B and OPT-125M. The best results are highlighted in bold for better comparison. We highlight the best results between FZOO and MeZO in bold to facilitate comparison.}

\resizebox{\textwidth}{!}{
  \setlength{\tabcolsep}{4pt}

  \begin{tabular}{lc*{14}{c}} 
    \toprule
    & \textbf{Model} & \textbf{Method} & \textbf{SST-2} & \textbf{RTE} & \textbf{CB} & \textbf{BoolQ} & \textbf{WSC} & \textbf{WIC} & \textbf{MultiRC} & \textbf{COPA} & \textbf{ReCoRD} & \textbf{SQuAD} & \textbf{DROP} & \textbf{Average}\\
    \midrule

    &OPT-2.7B &MeZO    &92.2 &58.8 &62.5 &64.0 &53.8 &54.2 &58.4 &76.0 &75.0 &77.6 &25.3 &63.4\\
    
    &OPT-2.7B &FZOO   &\textbf{93.6} &\textbf{69.0} &\textbf{62.5} &\textbf{70.8} &\textbf{57.7} &\textbf{57.5} &\textbf{65.5} &\textbf{83.0} &\textbf{76.2} &\textbf{81.8} & \textbf{26.5} &\textbf{67.6} \\ 
    \midrule

    &OPT-125M &MeZO    &81.8 &55.6 &67.9 &60.0 &60.6 &54.1 &58.7 &64.0 &\textbf{52.3} &44.1 &14.2 &55.8 \\

    &OPT-125M &FZOO   &\textbf{84.5} &\textbf{60.7} &\textbf{67.9} &\textbf{61.8} &\textbf{60.6} &\textbf{57.4 }&\textbf{61.1} &\textbf{64.0} &50.7 &\textbf{51.0} &\textbf{15.9} &\textbf{57.8}\\
    \bottomrule
  \end{tabular}%
} 
\label{tab:app_opt125m_2.7b}
\end{table*}

\section{Details about Memory Usage}
\label{app:memory_time}

Here we show the detailed numbers of memory profiling results Table \ref{tab:app_memory}. We did not turn on any advance memory-saving options, e.g., gradient checkpointing. We set the per-device batch size as 1 to test the minimum hardware requirement to run the model with specific optimization algorithms. We use Nvidia’s $nvidia-smi$ command to monitor the GPU memory usage.
\begin{table*}[h]
  \centering
  \caption{Memory usage on the MultiRC (average tokens=400) dataset. Results of ICL and full-parameter tuning are from MeZO\citep{mezo}.}
  \scalebox{0.82}{
    \begin{tabular}{lcccccc}
    \toprule
    &Method & \textbf{zero-shot/FZOO(FT)} & \textbf{HiZOO(FT) } & \textbf{Adam(Prefix)} & \textbf{ICL} & \textbf{Adam(FT)}\\
    \midrule
    &1.3B & 1xA100 (4GB) & 1xA100 (7GB) & 1xA100 (19GB) & 1xA100 (6GB)  & 1xA100 (27GB)\\
    &2.7B & 1xA100 (7GB) & 1xA100 (13GB) & 1xA100 (29GB) & 1xA100 (8GB)   & 1xA100 (55GB)\\
    &6.7B & 1xA100 (14GB) & 1xA100 (29GB) & 1xA100 (46GB) & 1xA100 (16GB)  & 2xA100 (156GB)\\
    &13B & 1xA100 (26GB) & 1xA100 (53GB) & 2xA100 (158GB) & 1xA100 (29GB)  & 4xA100 (316GB)\\
    &30B & 1xA100 (58GB) & 2xA100 (118GB) & 4xA100 (315GB) & 1xA100 (62GB)  & 8xA100 (633GB)\\
    &66B & 2xA100 (128GB) & 3xA100 (246GB) & 8xA100  & 2xA100 (134GB) & 16xA100\\

    \bottomrule
    \end{tabular}}
    
    \label{tab:app_memory}
\end{table*}%
\section{Details about Wallclock Time Efficiency}
\label{app:train_time}

As shown in Table~\ref{tab:app_train_time_old}, to avoid introducing additional overheads such as inter-GPU communication, results are measured on the same dataset (RoBERTa-large use SST-2 and OPT use COPA) and GPUs (80GB A100), with each result averaged over 100 steps. "BS" refers to batch size. For the relatively smaller RoBERTa-large model, we used a BS=64, while for other models, we used a BS=16.
\begin{table*}[htbp]
  \centering
  \vspace{-3pt}
  \caption{Wallclock time per step between Adam, MeZO and FZOO(N=8). }
  \vspace{4pt}
  \scalebox{1}{
    \begin{tabular}{lccc} 
    \toprule
    \textbf{Method} & \textbf{OPT-125M} & \textbf{RoBERTa-large} & \textbf{OPT-1.3B} \\
    \midrule
    Adam & 0.1982s(BS=16) & 0.3930s(BS=64) & 0.5814s(BS=16) \\
    MeZO & 0.1368s(BS=16) & 0.4305s(BS=64) & 0.7218s(BS=16) \\
    FZOO w/o parallel & 0.6941s(BS=16) & 1.0773s(BS=64) & 3.1925s(BS=16) \\
    FZOO  & 0.3835s(BS=16) & 0.6052s(BS=64) & 1.6628s(BS=16) \\
    \bottomrule
    \end{tabular}
  }
  
  \label{tab:app_train_time_old}

\end{table*}


\section{Details about Ablation Experiments}
\label{app:ablation}

\subsection{Influence of Number of Perturbation Directions Used in Each Update}
\label{app:smooth_scale}

We performed an ablation study on the perturbation batch size on the SST-2 dataset using the OPT-125M model. As shown in Table \ref{tab:batch_ablation}, we observed that a batch size of 8 shows the best average performance among all tested configurations. While some other settings perform well in certain cases, batch size 8 offers a good trade-off between stability and effectiveness, suggesting that a moderate number of perturbations per step benefits the zeroth-order optimization process.

\begin{table*}[htbp]

  \centering
  \caption{Ablations on OPT-125M using SST-2 dataset.}

    \scalebox{0.8}{
    \begin{tabular}{lcccccccc}
    \toprule
    Batch Size & \multicolumn{1}{c}{\textbf{(5e-5,1e-3)}} & \multicolumn{1}{c}{\textbf{(2e-4,5e-5)}} & \multicolumn{1}{c}{\textbf{(5e-4,1e-4)}}  & \multicolumn{1}{c}{\textbf{(1e-5,1e-4)}} & \multicolumn{1}{c}{\textbf{(1e-4,1e-3)}} & \multicolumn{1}{c}{\textbf{(5e-5,5e-5)}} & \multicolumn{1}{c}{\textbf{(5e-5,1e-4)}} & \multicolumn{1}{c}{\textbf{Average}} \\

    \midrule
    2  & 0.8154 & 0.4908 & 0.4908 & 0.4908 & 0.7901 & 0.8165 & 0.4908 & 0.6484\\

    4 & 0.8360 & 0.7798 & 0.6261 & \textbf{0.8349} & 0.8211 & 0.8417 & 0.8372 & 0.8026\\

    8  & \textbf{0.8429} & 0.8521 & 0.7626 & 0.8028 & \textbf{0.8440} & 0.8326 & \textbf{0.8417} & \textbf{0.8278}\\

    16 & 0.8326 & \textbf{0.8567} & 0.8211 & 0.7110 & \textbf{0.8440} & \textbf{0.8452} & 0.8349 & 0.8244 \\

    32  & 0.7982 & 0.8498 & \textbf{0.8429} & 0.5734 & 0.8268 & 0.8154 & 0.8119 & 0.7945\\

    \bottomrule
    \end{tabular}}

    \label{tab:batch_ablation}%
\end{table*}%
